\documentclass[lettersize,journal]{IEEEtran}
\usepackage{amsmath,amsfonts}
\usepackage{algorithmic}
\usepackage{algorithm}
\usepackage{array}
\usepackage[caption=false,font=normalsize,labelfont=sf,textfont=sf]{subfig}
\usepackage{textcomp}
\usepackage{stfloats}
\usepackage{url}
\usepackage{verbatim}
\usepackage{graphicx}
\usepackage{cite}
\usepackage{csquotes}
\usepackage{tabu}
\usepackage{multirow}
\usepackage{makecell}
\usepackage{colortbl}
\usepackage{enumitem}
\usepackage{pdfpages}
\usepackage{xcolor}

\newcommand{\revision}[1]{\textcolor{black}{#1}}
\newcommand{\secrevision}[1]{\textcolor{black}{#1}}
\newcommand\T{\rule{0pt}{2.9ex}}       
\newcommand\B{\rule[-1.2ex]{0pt}{0pt}} 
\newcommand{\concat}{\mathcal{}{||}}
\usepackage[pagebackref=true,breaklinks=true,letterpaper=true,colorlinks,bookmarks=false]{hyperref}
\usepackage{orcidlink}

\begin{document}

\title{HINT: High-quality INpainting Transformer with Mask-Aware Encoding and Enhanced Attention}

\author{Shuang Chen$^{\orcidlink{0000-0002-6879-7285}}$, Amir Atapour-Abarghouei$^{\orcidlink{0000-0002-4242-4579}}$, Hubert P. H. Shum$^{\orcidlink{0000-0001-5651-6039}\dag}$,~\IEEEmembership{Senior Member,~IEEE}
\thanks{Manuscript created august, 2023.}
\thanks{S. Chen, A. Atapour-Abarghouei and H. P. H. Shum are with Durham University, UK.  (e-mail: \{shuang.chen, amir.atapour-abarghouei, hubert.shum\}@durham.ac.uk).}
\thanks{$^{\dag}$Corresponding author: H. P. H. Shum}
}


\markboth{IEEE Transactions on Multimedia, 2024}%
{Chen \MakeLowercase{\textit{et al.}}: HINT: High-quality INPainting Transformer with Mask-Aware Encoding and Enhanced Attention}


\maketitle

\begin{abstract}
Existing image inpainting methods leverage convolution-based downsampling approaches to reduce spatial dimensions. This may result in information loss from corrupted images where the available information is inherently sparse, especially for the scenario of large missing regions.
Recent advances in self-attention mechanisms within transformers have led to significant improvements in many computer vision tasks including  inpainting. However, limited by the computational costs, existing methods cannot fully exploit the efficacy of long-range modelling capabilities of such models. 
In this paper, we propose an end-to-end High-quality INpainting Transformer, abbreviated as HINT, which consists of a novel mask-aware pixel-shuffle downsampling module (MPD) to preserve the visible information extracted from the corrupted image while maintaining the integrity of the information available for high-level inferences made within the model.
Moreover, we propose a Spatially-activated Channel Attention Layer (SCAL), an efficient self-attention mechanism interpreting spatial awareness to model the corrupted image at multiple scales. To further enhance the effectiveness of SCAL, motivated by recent advanced in speech recognition, we introduce a sandwich structure that places feed-forward networks before and after the SCAL module.
We demonstrate the superior performance of HINT compared to contemporary state-of-the-art models on four datasets, CelebA, CelebA-HQ, Places2, and Dunhuang.

\end{abstract}


\section{Introduction}
\IEEEPARstart{I}{mage} inpainting is a computer vision task that aims to reconstruct an image based on the visible pixels of a damaged or corrupted image with missing regions. Its applications span across image processing and computer vision tasks such as photo editing~\cite{jo2019sc}, 
objective removal~\cite{wei2019shadow} and depth completion~\cite{atapour2019dealing}.

Image inpainting has been greatly benefited by modern learning-based techniques \cite{pathak2016context,iizuka2017globally,liu2018image,yu2019free,cao2021learning,yu2018generative,guo2021image,wang2019musical,sun2022learning}, even though it has existed long before the widespread use of deep learning  \cite{sridevi2019image,atapour2016back,barnes2009patchmatch}.
Many existing image inpainting methods \cite{zhang2022w, zhang2022pluralistic} with Convolutional Neural Networks (CNN) commonly use an encoder-decoder architecture, which down-sample the corrupted image to a hidden latent space, and then up-sample it to produce a restored image with comparable semantics and structure to the original~\cite{pathak2016context,iizuka2017globally}.
These CNN-based methods typically use local convolutional filters, which possess a limited receptive field and solely capture information within a restricted local region. These methods suffer from the limitation to capture long-range spatial relations between distant image regions, thus compromising their effectiveness in image inpainting~\cite{pathak2016context,nazeri2019edgeconnect,liu2018image}.
To address this challenge, ~\cite{uddin2020global, zheng2022bridging} incorporate spatial self-attention into the network, which involves calculating a self-attention map in the deep latent space to capture long-range dependencies between feature elements. Recent work attempted to introduce transformer architectures to image inpainting ~\cite{wan2021high,zheng2022bridging,li2022mat}, which enable global long-range modeling of images that have been down-sampled or partitioned into patches, resulting in improved performance.

\begin{figure}[t] \centering
    \makebox[0.01\textwidth]{}
    \\
    \raisebox{0.1\height}{\makebox[0.01\textwidth]{\rotatebox{90}{\makecell{\scriptsize CelebA-HQ}}}}
    \includegraphics[width=0.108\textwidth]{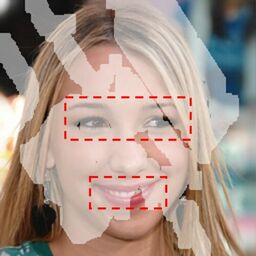}
    \includegraphics[width=0.108\textwidth]{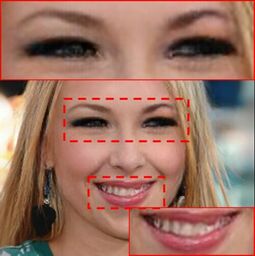}
    \includegraphics[width=0.108\textwidth]{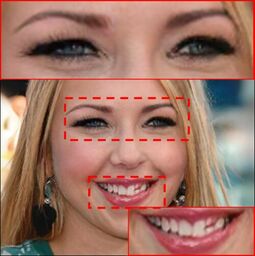}
    \includegraphics[width=0.108\textwidth]{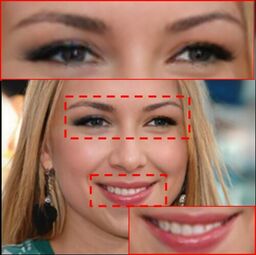}
    \\
    \vspace{-0.5em}
    \makebox[0.108\textwidth]{\scriptsize Masked input}
    \makebox[0.115\textwidth]{\scriptsize LaMa~\cite{suvorov2022resolution}}
    \makebox[0.11\textwidth]{\scriptsize MAT~\cite{li2022mat}}
    \makebox[0.1\textwidth]{\scriptsize Ours}
    \\
    \raisebox{0.5\height}{\makebox[0.01\textwidth]{\rotatebox{90}{\makecell{\scriptsize Places2}}}}
    \includegraphics[width=0.108\textwidth]{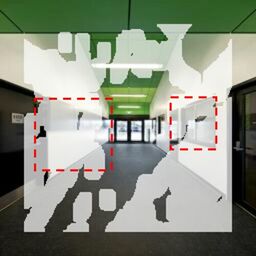}
    \includegraphics[width=0.108\textwidth]{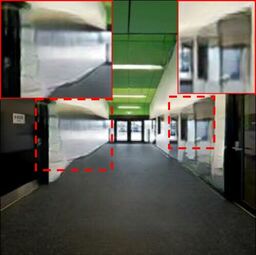}
    \includegraphics[width=0.108\textwidth]{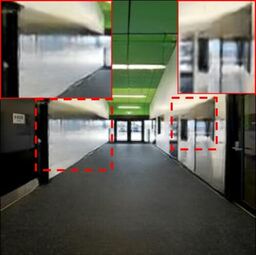}
    \includegraphics[width=0.108\textwidth]{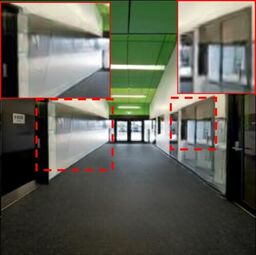}
    \\
    \vspace{-0.5em}
    \makebox[0.108\textwidth]{\scriptsize Masked input}
    \makebox[0.115\textwidth]{\scriptsize MISF~\cite{li2022misf}}
    \makebox[0.11\textwidth]{\scriptsize LaMa~\cite{suvorov2022resolution}}
    \makebox[0.1\textwidth]{\scriptsize Ours}
    \\
    \raisebox{0.3\height}{\makebox[0.01\textwidth]{\rotatebox{90}{\makecell{\scriptsize Dunhuang}}}}
    \includegraphics[width=0.108\textwidth]{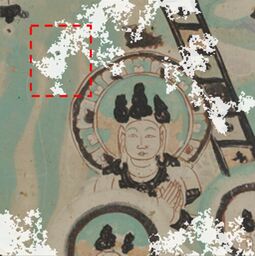}
    \includegraphics[width=0.108\textwidth]{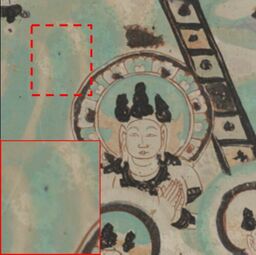}
    \includegraphics[width=0.108\textwidth]{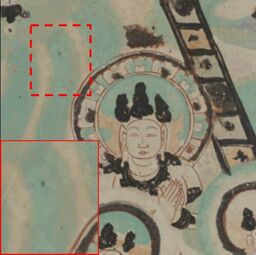}
    \includegraphics[width=0.108\textwidth]{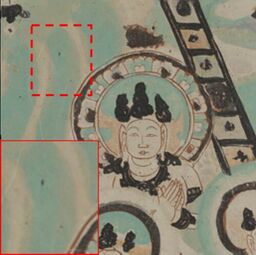}
    \\
    \vspace{-0.5em}
    \makebox[0.108\textwidth]{\scriptsize Masked input}
    \makebox[0.115\textwidth]{\scriptsize MISF~\cite{li2022misf}}
    \makebox[0.11\textwidth]{\scriptsize JPGNet~\cite{guo2021jpgnet}}
    \makebox[0.1\textwidth]{\scriptsize Our HINT}
    \\
    \raisebox{0.3\height}{\makebox[0.01\textwidth]{\rotatebox{90}{\makecell{\scriptsize Example 1}}}}
    \includegraphics[width=0.22\textwidth]{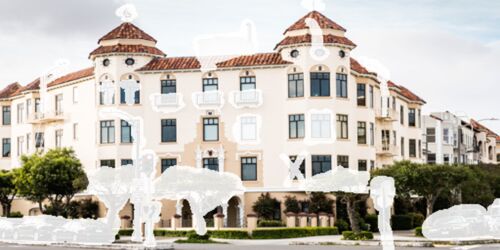}
    \includegraphics[width=0.22\textwidth]{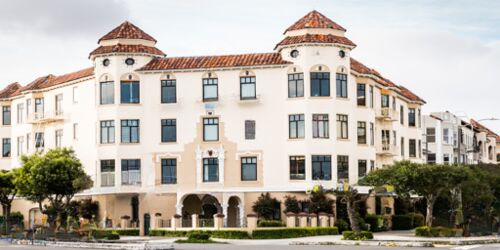}\\
    \raisebox{0.3\height}{\makebox[0.01\textwidth]{\rotatebox{90}{\makecell{\scriptsize Example 2}}}}
    \includegraphics[width=0.22\textwidth]{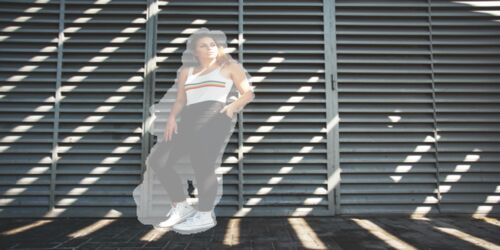}
    \includegraphics[width=0.22\textwidth]
    {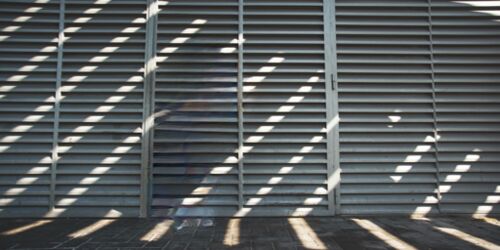}
    \\
    \vspace{-0.5em}
    \makebox[0.22\textwidth]{\footnotesize Masked Input}
    \makebox[0.22\textwidth]{\footnotesize Ours}

    \vspace{-0.5em}

    \caption{Comparisons with the state of the art~\cite{suvorov2022resolution,li2022mat,li2022misf} on different datasets~\cite{karras2017progressive,zhou2017places,yu2019dunhuang} with large masks (shown in white areas). Red boxes highlight major differences. The bottom two examples are from unseen real-world high-resolution images.} 
    \vspace{-1.2em}
    \label{fig:teaser}
\end{figure}

A significant challenge hindering image inpainting is effectively modeling the valid information within visible regions, which is crucial for reconstructing semantically coherent and texture-consistent details in the missing regions. This is particularly noticeable in large masked regions, where the valid information is limited. 
Existing methods that utilise convolutional layers for downsampling come with the inherent drawback of information loss~\cite{zhao2017random}, attributed to the reduction of feature size from filters and downsampling. 
Given its capability to preserve 
input information, pixel-shuffle down-sample 
is widely used in image denoising~\cite{a2021beyond}, image deraining~\cite{yue2021semi} and image super-resolution~\cite{wang2021learning}.
It periodically rearranges the elements of the input into an output scaled by the sample stride.
\revision{However, its effectiveness depends on the assumption that the sample stride is small enough to avoid disrupting the noise distribution~\cite{zhou2020awgn}. This holds only for a relatively independent distribution of raindrops and noise, and is not suitable for image inpainting with irregular and variable-size masks. Simply using conventional Pixel-shuffle Down-sampling (PD) \cite{a2021beyond,yue2021semi,wang2021learning} for corrupted image would lead to the problem of pixel drifting, which is shown in Fig.~\ref{fig:MPD} (upper branch). The pixel drifting happens in $\hat{X}$. After the feature ${X}'$ is downsampled, the position of the masked regions (white elements) becomes inconsistent across channels, causing the visible area to be misaligned in the channel, disrupting subsequent feature extraction processes within the model, thus affecting the accurate modelling of the valid information from the visible regions of the input image.}

Another challenge in applying spatial self-attention in CNN-based models is its significant computational expense. Considering this, spatial self-attention is typically only employed on low-resolution representations~\cite{uddin2020global, zheng2022bridging}. 
While transformer-based methods~\cite{wan2021high,li2022mat} employ multiple spatial self-attention blocks to model long-range dependencies. However, the quadratic computational complexity limits their wider applicability. To address this, the prevalent compromise involves down-sampling~\cite{li2022mat} or reducing the resolution~\cite{wan2021high} of the input image prior to being passed through the transformer. 
However, this strategy leads to information loss from the input images through the model, which is detrimental to image inpainting where visible information is already limited. This loss subsequently results in the degradation of fine-grained features. As long-range dependencies are modelled over these degraded features, the reconstructed output suffer from blurring artefacts and vague structures. \cite{wan2021high,zheng2022bridging} introduce extra refinement networks to improve image quality after getting coarse completed images, rather than recovering high-quality results directly. 
\revision{The method in \cite{zamir2022restormer} replaces spatial self-attention with channel self-attention to reduce computational complexity. Although channel self-attention gains linear computational complexity, it completely loses spatial awareness. This makes it possible to highlight \enquote{what} the salient features are but cannot discern \enquote{where} the spatially important regions are, which is essential as visible regions often exhibit complex and irregular shapes, especially with large irregular masks. Some existing works~\cite{69,70,71} attempt to address the spatial awareness loss by incorporating spatial self-attention back to the channel self-attention, but at a cost of significant increases in computation.}


To address these common challenges currently restricting progress in the existing literature, we present a novel High-quality INpainting Transformer (HINT) for image inpainting, which enables efficient multiscale modeling of the global context while minimising the loss of valid information. 
Specifically, we propose a tailor-made pixel-shuffle down-sampling (MPD) module for image inpainting to reduce information loss and maintain the consistency of data. 
\revision{To enhance the representation learning capabilities of our model, we develop a Spatially-activated Channel Attention Layer (SCAL) to blend information in both the channel and spatial dimensions. Unlike these existing methods~\cite{69,70,71}, the innovation of SCAL lies in its minimalistic and efficient design, only utilising convolutional layers to retrain spatial awareness, thereby mitigating the significant computational cost, which is a major issue in the field. This enhanced self-attention module plays the predominant role in HINT and build HINT as a transformer-based model.}
To further improve the effectiveness of SCAL with limited parameters, we propose a module known as the ``Sandwich'', sandwiching the proposed SCAL between two feed-forward networks (FFNs) for each transformer block. This structure results in better performance compared to alternative designs with the same number of network parameters. 



Comparative experiments show that HINT outperforms state-of-the-art image inpainting approaches  
(Fig.~\ref{fig:comparison_CelebAHQ&Places2}) across four datasets, i.e., CelebA~\cite{liu2015deep}, CelebA-HQ~\cite{karras2017progressive}, Places2~\cite{zhou2017places} and Dunhuang challenge~\cite{yu2019dunhuang}. We also perform ablation experiments to demonstrate the contribution of proposed components in HINT.

Our source code is openly released at \url{https://github.com/ChrisChen1023/HINT}.

Our major contributions are as follows:
\begin{itemize}[noitemsep,topsep=0pt,labelindent=0cm,leftmargin=0.4cm]
\item We propose HINT, an end-to-end transformer-based architecture for image inpainting that takes advantage of multi-scale feature- and spatial-level representations as well as pixel-level visual information.
\item
We propose a plug-and-play mask-aware pixel-shuffle down-sampling (MPD) module to preserve useful information while keeping irregular masks consistent during downsampling (Section \ref{sec:methodology:mpd}).
\item
\revision{We propose a Spatially-activated Channel Attention Layer (SCAL) using self-attention and convolutional attention to sequentially refine features at the channel and spatial dimensions. We further design an improved sandwich-shaped transformer block to boost the efficacy of the proposed SCAL (Section \ref{sec:methodology:transformer_body})}.

\end{itemize}

\section{Related Work}
We consider prior work within two distinct \revision{areas}:
image inpainting (Section~\ref{sec:related_work:inpainting}), and visual transformers (Section~\ref{sec:related_work:transformers}), which have gained prominence as effective techniques for addressing the image inpainting task.
\subsection{Image Inpainting}
\label{sec:related_work:inpainting}
Image inpainting predates learning-based techniques and the literature on image completion based on conventional strategies is extensive. Diffusion-based approaches complete minor and narrow stretches using neighbouring visible pixels \cite{sridevi2019image}. Exemplar-based methods infer missing regions with plausible edge information based on other patches from background or external data \cite{atapour2016back,barnes2009patchmatch}. 
However, despite their ability to plausibly reconstruct images with small and constrained missing regions, these methods \revision{are not fully capable of} generating innovative features if they do not already exist in the known regions of images.

Compared with traditional methods, learning-based methods have achieved great success in inpainting, especially when it comes to generating new contextually sound content for large missing regions. \cite{pathak2016context} proposed a parametric framework for image inpainting based on an encoder-decoder architecture taking advantage of a Generative Adversarial Network (GAN)~\cite{goodfellow2014generative}. Subsequently, numerous GAN-based methods emerged to offer improved inpainting quality \cite{liu2018image,yu2019free,cao2021learning,li2020recurrent,peng2021generating,guo2021image, nazeri2019edgeconnect, wu2021deep, chen2023inclg} using better training strategies. 

\cite{iizuka2017globally} use two discriminators to calculate both global and local adversarial losses. \cite{yu2020region} propose region-wise normalisation for missing and visible areas. Partial~\cite{liu2018image} and gated convolutions ~\cite{yu2019free,cao2021learning} are introduced to handle the irregular masks by improving the convolution operation~\cite{li2020recurrent, peng2021generating} to efficiently extract valid information for inpainting. ~\cite{yu2018generative} propose contextual attention to facilitate the matching of feature patches across distant spatial locations. \revision{Building on this, ~\cite{guo2021image,wang2019musical} extended \cite{yu2018generative} by incorporating a multi-scale patch size to further improve its efficiency. \cite{76} introduces fourier convolution-based encoder for image inpainting to avoid generating invalid feature inside the missing regions.}
These strategies all aim to explore how to effectively extract valid information from known regions for hole-filling, but they still suffer from the information loss caused by convolutional downsampling.


\subsection{Visual Transformers}
\label{sec:related_work:transformers}
The notable success of Transformers~\cite{vaswani2017attention} in natural language processing has recently prompted research into their applicability in computer vision~\cite{liu2021swin,dosovitskiy2020image}. Driven by this, efforts were focused towards applying transformers to image inpainting ~\cite{yu2021diverse,wan2021high,zheng2022bridging, zhang2023mutual,li2022mat,deng2021learning}. 
However, spatial-based self-attention incurs an expensive computational cost. To reduce computation, \cite{yu2021diverse,wan2021high} down-sample the input image into a lower resolution.~\cite{zheng2022bridging,li2022mat,deng2021learning} calculate the spatial self-attention after encoding the input image into low-resolution features. Nonetheless, these approaches fail to change the quadratic complexity of spatial self-attention, which restricts its applicability to high-frequency features.

Swin Transformer~\cite{liu2021swin} reduces the computational complexity to linearity. However, the shifted-window design splits the local neighbourhood context of the visible and missing area, and thus is not ideally suited for inpainting. ~\cite{zamir2022restormer} propose utilising channel-wise self-attention in multi-scale representation with linear complexity for image reconstruction. Its variant~\cite{deng2022t} demonstrates the applicability in image inpainting. Nevertheless, both of these models omit spatial attention that is vital in delivering high-quality and contextually sound results.
In contrast, our model integrates multiscale channel and spatial attention in an efficient manner, thus resolving the issue that prior work has struggled with \cite{wan2021high,zheng2022bridging,li2022mat}.

\begin{figure*}
\begin{center}
\centerline{\includegraphics[scale=2]{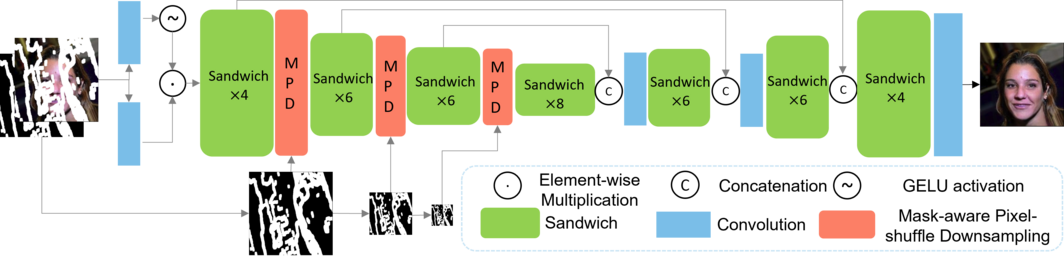}}
\end{center}
    \vspace{-1.5em}
   \caption{The overview of the proposed framework, which is built with a gated embedding block, with multiple stacked \enquote{sandwiches} in different levels. The \enquote{sandwich} is described in Sec.~\ref{sec:methodology:transformer_body:sandwich}, the MPD is described in Sec.~\ref{sec:methodology:mpd} 
   }
\label{overview}
\end{figure*}

\section{HINT: High-quality INpainting Transformer}
\revision{Formally, the problem is formulated as follows: the input image, ${I}_{input}$, is obtained by concatenating masked image, ${I}_{M}={I}\odot{M}$, and the mask, ${M}$. The input image, ${I}_{input}$, is then processed by our proposed HINT model and a semantically accurate output image, ${I}_{C}$, will be generated. The whole formulation is denoted as: ${I}_{C}=HINT({I}_{input})$.}

We present our transformer-based HINT approach to image inpainting, which takes advantage of our novel Mask-aware Pixel-shuffle Down-sampling (MPD) to solve the information loss issue during downsampling and further enhance the use of valid information from known areas. Within the architecture, we propose a Spatially-activated Channel Attention Layer (SCAL), which aims to handle spatial awareness while maintaining efficiency within the transformer block. The SCAL is encapsulated between two feed-forward networks, forming a sandwich-shaped transformer block, henceforth referred to as \enquote{\textit{Sandwich}}. This design enables the effective extraction of long-range dependencies while preserving the smooth and coherent flow of valid information through the model.



\subsection{The Overall Pipeline}
Overall, as seen in Fig. \ref{overview},  HINT consists of an end-to-end network with a gated embedding layer to selectively extract features, followed by a transformer body for modelling long-range correlations, and a projection layer to generate the output. Specifically, we insert a gating mechanism ~\cite{yu2019free} into the embedding layer serving as a feature extractor, achieved by using two parallel paths of vanilla convolutions with one path activated by a GELU non-linearity~\cite{hendrycks2016gaussian} to dynamically embed the finer-grained features, leading to stronger representation learning and better optimisation ~\cite{xiao2021early}. 
The transformer body is an encoder-decoder architecture comprising multiple transformer blocks. The encoder consists of the first three blocks, each followed by an MPD layer to mitigate incoherence in invalid locations, while the final three blocks with conventional pixel shuffle upsampling form the decoder. Mirrored blocks are connected via skip connections to preserve shared features learned within the encoder. At the end, a convolutional layer is used to project the decoded features to the final output. 




\subsection{Mask-aware Pixel-shuffle Down-sampling}
\label{sec:methodology:mpd}

\begin{figure}[t]
\begin{center}
\centerline{\includegraphics[scale=2]{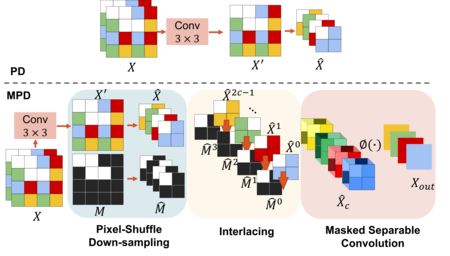}}
\end{center}
    \vspace{-1.5em}
   \caption{\revision{The comparison of Pixel-shuffle Down-sampling (PD, upper) and the proposed Mask-aware Pixel-shuffle Down-sampling (MPD, lower). Ours proposed MPD, with one $3 \times 3$ convolution, a conventional PD, interlacing (concatenation of feature and mask slices), and a masked-separable convolution. Invalid pixel drifting happens in $\hat{X}$. After the feature ${X}'$ is downsampled, the masked position becomes inconsistent across channels.}}
\label{fig:MPD}
\end{figure}

Conventional Pixel-shuffling Down-sampling (PD) is the inverse operation of Pixel-shuffle~\cite{shi2016real}. It periodically rearranges the input ${{T}_{in} \in \mathbb{R}^{{H}\times {W}\times {C}}}$ into ${{T}_{out}\in\mathbb{R}^{\frac{H}{s} \times \frac{W}{s} \times  {s}^{2}{C}}}$
for downsampling with $s$ being the scale factor to denote the sample stride. PD can effectively preserve the input information, which is desirable for inpainting, particularly for reconstructing high-quality images. However, as PD uses non-overlapping sampling with stride $s$ to generate mosaics from the image ~\cite{shi2016real}, the consistency of missing pixel locations can be disrupted during the down-sampling, as shown in Fig.~\ref{fig:MPD}, making it unsuitable for image inpainting.


We propose a Mask-aware Pixel-shuffle Down-sampling (MPD) module, which is a novel down-sampling approach specifically tailored for image inpainting. It resolves the issue of positional drift of masked pixels that occurs during the process of conventional PD. Furthermore, in contrast to convolution-downsampling, MPD preserves all valid information, thereby minimising information loss. Apart from inpainting, this module can be plugged into any other problem that involves masking, such as any that might use image segmentation labels masks as their input.


Given the features ${X}\in \mathbb{R}^{H \times W \times C}$ and mask ${M}\in \mathbb{R}^{H \times W \times 1}$, we first project $X$ into $X'$ with half the channels but the same size ~\cite{shi2016real}, utilising a $3 \times 3$ convolution operator $h(\cdot)$, and perform PD on both ${X}'$ and ${M}$:
\begin{equation}
\begin{gathered}
\hat{M}=PD(M), \hat{X}=PD(h({X})).
\end{gathered}
\label{equ:MPD}
\end{equation}


As shown in Fig.~\ref{fig:MPD}, the positions of the missing pixels in $\hat{X}$ drift and are discontinuous across channels while each channel of $\hat{M}$ sequentially indicates the positions of valid and invalid pixels in $\hat{X}$. To enforce $\hat{M}$ to act on the corresponding channel accurately, we intersperse and concatenate the sliced $\hat{X}$ and $\hat{M}$ across the channel, obtaining $\hat{X}_{c}\in\mathbb{R}^{\frac{H}{2} \times \frac{W}{2} \times 2 C}$.
\begin{equation}
\begin{gathered}
\hat{M^0},\hat{M^1},\hat{M^2},\hat{M^3} = Slice (\hat{M}),\\
\hat{X^0},\hat{X^1},\hat{X^2},\dots,\hat{X}^{2C-1} = Slice (\hat{X}),
\end{gathered}
\label{slice}
\end{equation}
\vspace{-1em}
\begin{equation}
\begin{aligned}
\hat{X}_{c} = &(\hat{X^0}\concat\hat{M^0})\concat\dots \concat(\hat{X^i}\concat\hat{M}^{(i+4)\%4})\concat\dots\\ 
&\concat(\hat{X}^{2C-1}\concat\hat{M^3}),
\end{aligned}
\label{MPD}
\end{equation}
where $Slice(\cdot)$ is a channel-wise slice, $\concat$ is channel-wise concatenation, and $\%$ denotes the modulo operator. \revision{Thus, each feature has a paired mask as an indicator. In the end, we exploit a separable convolutional layer~\cite{chollet2017xception}, denoted as $\phi(\cdot)$, to encode pairs of features and masks, aiming to learn the correct local priors from the features indicated by the shuffled mask, and forcing the encoder to accurately model the valid information within the visible regions. The output is formulated as:}
\begin{equation}
\begin{aligned}
X_{out} = \phi(\hat{X}_{c}).
\end{aligned}
\end{equation}

\subsection{The Transformer Body}
\label{sec:methodology:transformer_body}
Each of the seven transformer blocks stacks multiple \textit{sandwiches} encapsulating the proposed SCAL for local-global representation learning, working with MPD to down-sample the features and control data flow consistency (Fig. \ref{overview}).

\subsubsection{Spatially-activated Channel Attention Layer}
\label{sec:methodology:transformer_body:scal}
We propose a Spatially-activated Channel Attention Layer (SCAL) to strengthen the model to capture inter-channel dependencies while preserving spatial awareness. 
Channel self-attention~\cite{fu2019dual} is computationally viable for high-resolution features due to its linear time and memory complexity growth with channel depth. 
However, it fails to account for \enquote{where} the important information is across the entire spatial position, thus ignoring the relationship between feature patches. This is very important for image inpainting as the global context in the valid regions within each image can be distinct and irregularly shaped, as defined by the irregular mask ${M}$.

To alleviate this issue, we improve the concept of transposed attention~\cite{zamir2022restormer} by introducing a convolution-attention branch to capture the attention matrix of spatial locations. 
This enables HINT to effectively model long-range dependencies in the channel dimension, while attending to spatial locations where features should be emphasised. Unlike alternative approaches~\cite{liu2021swin,dosovitskiy2020image,li2022mat,zheng2022bridging,wan2021high}, SCAL does not increase the computational cost quadratically with input resolution, making it feasible for multi-scale context modelling.

\begin{figure}[t]
\begin{center}
\centerline{\includegraphics[scale=2]{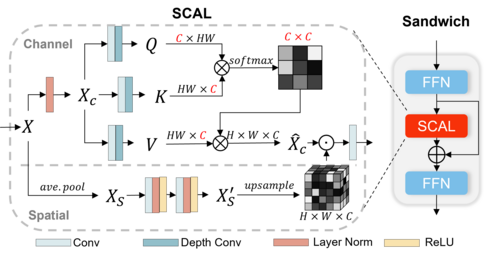}}
\end{center}
    \vspace{-1.5em}
   \caption{\revision{``Sandwich'' (right) and ``Spatially-activated Channel Attention Layer'' (left). \enquote{$\bigoplus$},\enquote{$\bigotimes$}, and \enquote{$\bigodot$} denote the element-wise sum, matrix multiplication, and element-wise multiplication, respectively.}
   }
\label{sandwich}
\end{figure}

As shown in Fig.~\ref{sandwich}, SCAL contains two branches. Given input feature ${X}$, the channel self-attention branch is:
\begin{equation}
\begin{gathered}
{X}_{c} = LN(X),\\
\hat{X}_{c}=\left(W_{d3}^V W_1^V {X}_{c}\right) \cdot \operatorname{Attc}({X}_{c}), \\
\operatorname{Attc}({X}_{c})=\varphi\left(\frac{W_{d3}^Q W_1^Q {X}_{c} \cdot\left(W_{d3}^K W_1^K {X}_{c}\right)^T}{\gamma}\right),
\end{gathered}
\label{eq:self-att}
\end{equation}
where \textit{LN} denotes layer normalisation, $\gamma$ is a learnable parameter to scale the dot product of \text{key} and \text{query}, $W_1$ is the linear projection and $W_{d3}$ is the $3 \times 3$ depth-wise convolution, $\operatorname{Attc}(\cdot)$ represents the function to calculate the channel attention map, and $\varphi$ is a softmax layer. In the spatial branch, we first downsample the input features $X$ but not fully squeeze, via average pooling to preserve global spatial information. Subsequently, two $3\times3$ convolutions serve as attention descriptors followed by an upsampling process, generating a soft global attention matrix, $\alpha = \operatorname{Atts}(X)$, which is used to reweight the output obtained through channel attention:
\begin{equation}
\begin{gathered}
\operatorname{Atts}(X)=\operatorname{Up}({f(g(\operatorname{AP}(X)))}),
\end{gathered}
\end{equation}
where $\operatorname{AP}$ is an average pooling layer, $\operatorname{Up}$ is upsampling. $f(\cdot)$ and $g(\cdot)$ are two similar convolution blocks, one of which contains a $3\times3$ convolutional layer, a normalisation layer, and a ReLU layer~\cite{fukushima1975cognitron}. $\operatorname{Atts}(\cdot)$ represents the function to calculate the spatial attention map.
As depicted in Fig.~\ref{sandwich}, the attention matrix $\alpha$ modulates the output of the channel branch $\hat{X}_c$ through point-wise multiplication. Subsequently, the mapping function $\theta(\cdot)$ is a projection layer performed via of a $1\times1$ convolution.
The complete representation of the SCAL is:
\begin{equation}
\begin{gathered}
SCAL(X)=\theta(\hat{X_c}\odot\operatorname{Atts}(X)).
\end{gathered}
\label{sq:scal}
\end{equation}

\subsubsection{Sandwich-shaped Transformer Block}
\label{sec:methodology:transformer_body:sandwich}
\revision{Image inpainting presents a significant challenge: the network must effectively learn from limited context to reconstruct complete images. This task is particularly daunting when faced with irregularly shaped masks, which complicate feature extraction, especially in areas with extensive missing information. This process of masking in image inpainting bears a notable resemblance to the masking of audio spectrograms in speech recognition for data augmentation purposes, as seen in techniques like SpecAugment~\cite{park2019specaugment,park2020specaugment}. The Conformer~\cite{gulati2020conformer}, with its innovative \enquote{FFN-Attention-Conv-FFN} architecture, demonstrates remarkable efficiency in speech recognition by using augmented, masked spectrograms as inputs. We hypothesise that such structures are equally effective for image inpainting, since their inputs are also incomplete and insufficient, highlighting a common challenge in both fields that may benefit from similar architectural solutions.}


Therefore, to boost the effectiveness of our attention layer, we propose a sandwich-shaped transformer block with an FFN-Attention-FFN structure. This first FFN serves as a filter, extracting more essential features for the following attention layer to capture long-distance dependencies (see Section~\ref{ablation-sandwich} for validations). 
Unlike~\cite{gulati2020conformer}, we remove the convolutional layer in the middle,
and enhance the two FFNs with depth-wise convolutions with a gate mechanism~\cite{zamir2022restormer}. This is because FFN integrating depth-wise convolution captures local information from every channel, which helps the model learn a more comprehensive and informative feature representation with fewer parameters~\cite{chollet2017xception}. Also, the gating strategy selectively filters and modulates the information flow according to the importance of each feature to the final high-quality output, thereby reducing irrelevant information and highlighting the most salient input features for representation learning. Given an input ${X} \in \mathbb{R}^{H \times W \times C}$, our sandwich is formulated as:
\begin{equation}
{X}_{\text{out}} = FFN(SCAL(FFN({X}))).
\label{stb}
\end{equation}

\subsection{Loss Functions}
To obtain high-quality inpainting results, we follow the established literature ~\cite{nazeri2019edgeconnect,li2022misf} to develop multiple loss components, including an $\mathcal{L}_1$ loss to enforce a contextually sound reconstruction, style loss $\mathcal{L}_{\text {style }}$ to measure the difference in style, perceptual loss $\mathcal{L}_{\text {perc}}$ to compare the high-level perceptual features extracted from a pre-trained network, and an adversarial loss $\mathcal{L}_{\text {adv}}$ to improve overall output quality. The final loss function is thus denoted as:
\begin{equation}
\begin{aligned}
\mathcal{L}_{total}(\hat{\mathbf{I}}, \mathbf{I_{gt}})=&\lambda_1 \mathcal{L}_1+\lambda_2 \mathcal{L}_{\text {style}}+\lambda_3 \mathcal{L}_{\text {perc}}\\
&+\lambda_4 \mathcal{L}_{\text {adv}},
\end{aligned}
\label{att2}
\end{equation}
where the weighting coefficients $\lambda_1$ = 1, $\lambda_2$ = 250, $\lambda_3$ = 0.1, $\lambda_4$ = 0.001 were chosen based on the parameter analysis (see Section~\ref{Parameter_tuning}).

\begin{figure*}[t] \centering
    \includegraphics[width=0.092\textwidth]{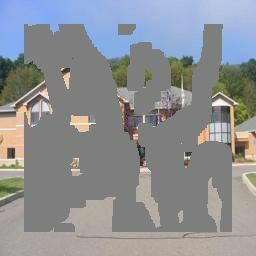}
    \includegraphics[width=0.092\textwidth]{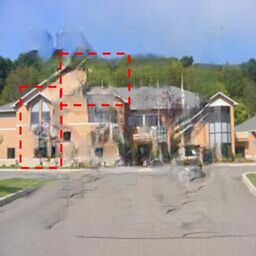}
    \includegraphics[width=0.092\textwidth]{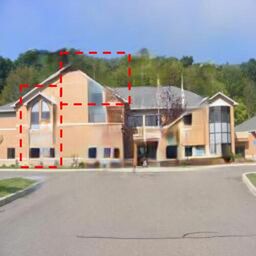}
    \includegraphics[width=0.092\textwidth]{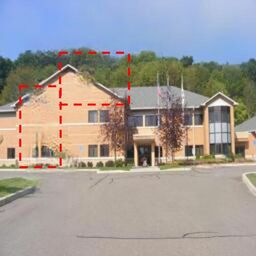}
    \includegraphics[width=0.092\textwidth]{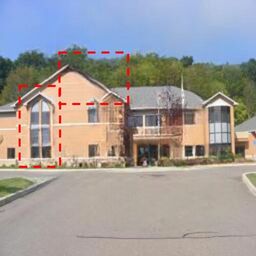}
    \includegraphics[width=0.092\textwidth]{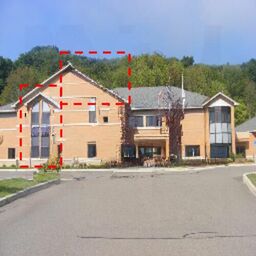}
    \includegraphics[width=0.092\textwidth]{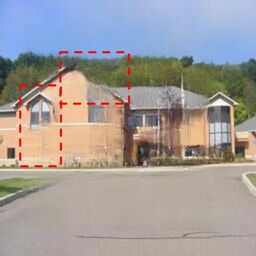}
    \includegraphics[width=0.092\textwidth]{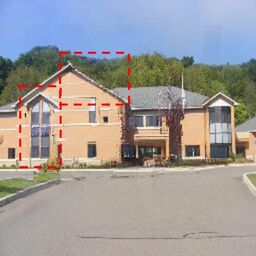}
    \includegraphics[width=0.092\textwidth]{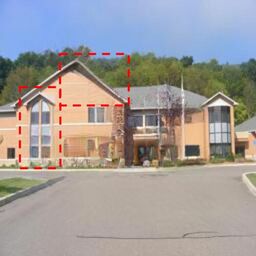}
    \includegraphics[width=0.092\textwidth]{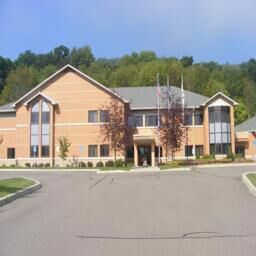}
    \\
    \includegraphics[width=0.092\textwidth]{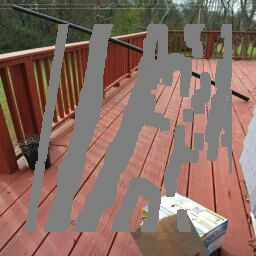}
    \includegraphics[width=0.092\textwidth]{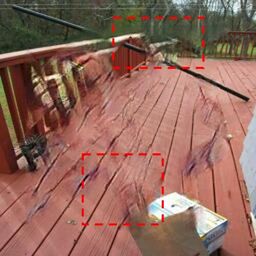}
    \includegraphics[width=0.092\textwidth]{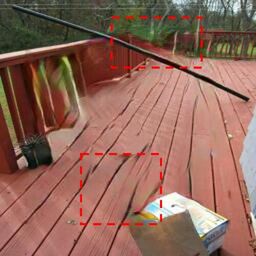}
    \includegraphics[width=0.092\textwidth]{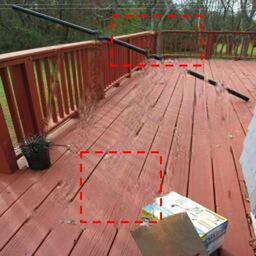}
    \includegraphics[width=0.092\textwidth]{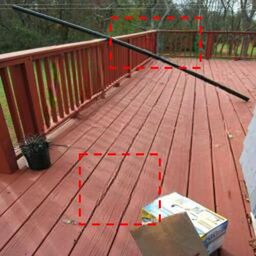}
    \includegraphics[width=0.092\textwidth]{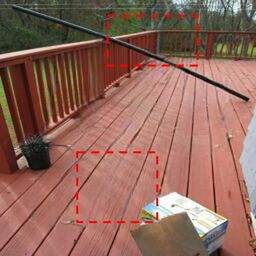}
    \includegraphics[width=0.092\textwidth]{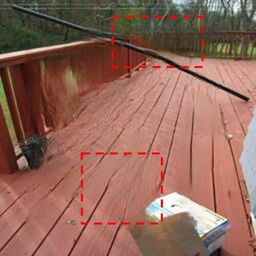}
    \includegraphics[width=0.092\textwidth]{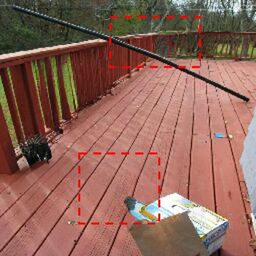}
    \includegraphics[width=0.092\textwidth]{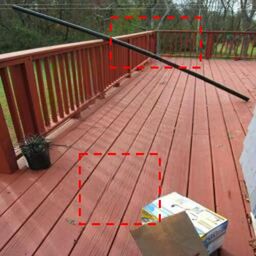}
    \includegraphics[width=0.092\textwidth]{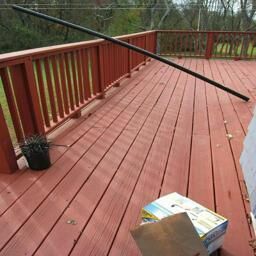}
    \\
    \includegraphics[width=0.092\textwidth]{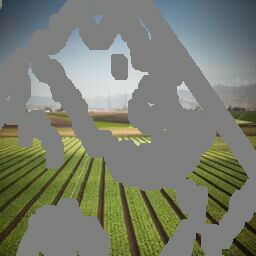}
    \includegraphics[width=0.092\textwidth]{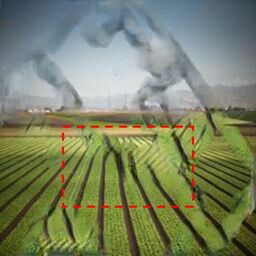}
    \includegraphics[width=0.092\textwidth]{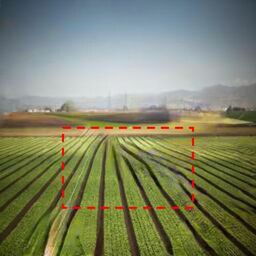}
    \includegraphics[width=0.092\textwidth]{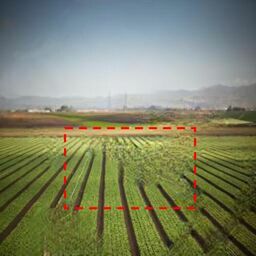}
    \includegraphics[width=0.092\textwidth]{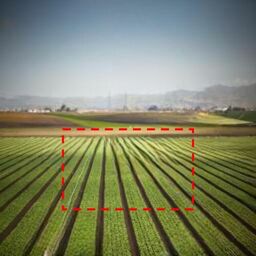}
    \includegraphics[width=0.092\textwidth]{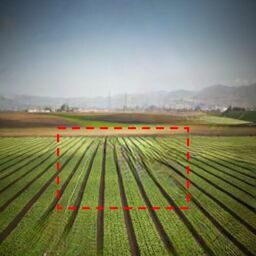}
    \includegraphics[width=0.092\textwidth]{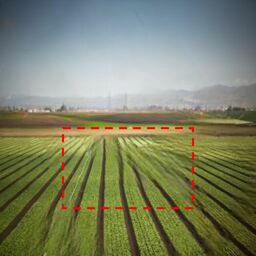}
    \includegraphics[width=0.092\textwidth]{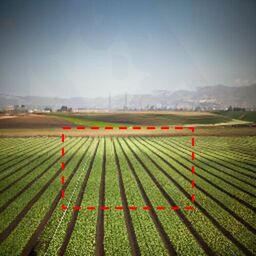}
    \includegraphics[width=0.092\textwidth]{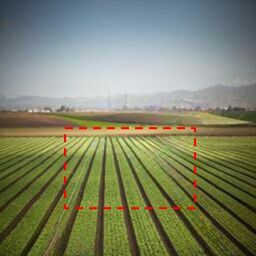}
    \includegraphics[width=0.092\textwidth]{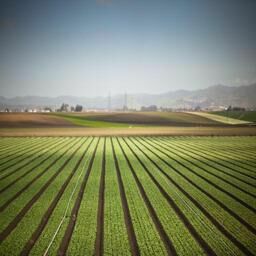}
    \\
    \includegraphics[width=0.092\textwidth]{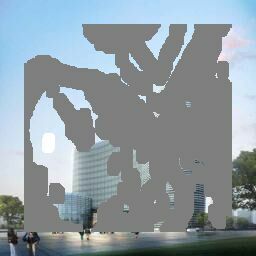}
    \includegraphics[width=0.092\textwidth]{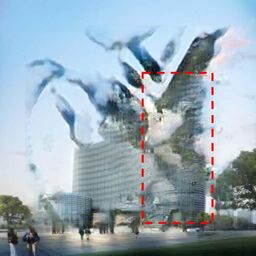}
    \includegraphics[width=0.092\textwidth]{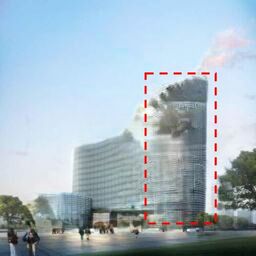}
    \includegraphics[width=0.092\textwidth]{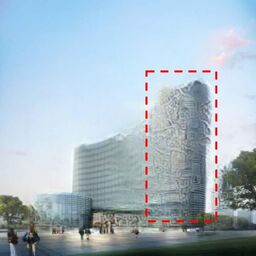}
    \includegraphics[width=0.092\textwidth]{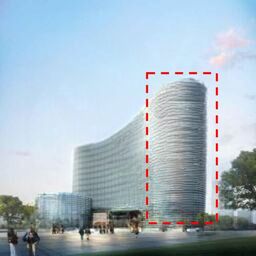}
    \includegraphics[width=0.092\textwidth]{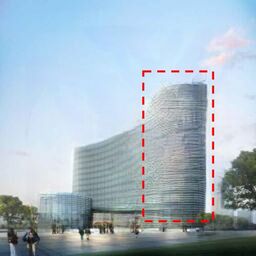}
    \includegraphics[width=0.092\textwidth]{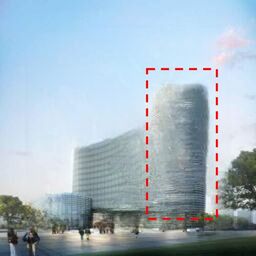}
    \includegraphics[width=0.092\textwidth]{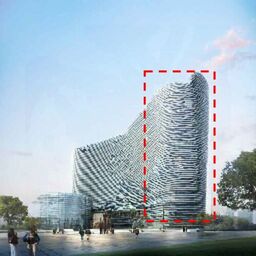}
    \includegraphics[width=0.092\textwidth]{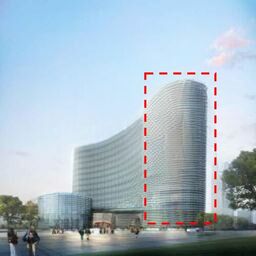}
    \includegraphics[width=0.092\textwidth]{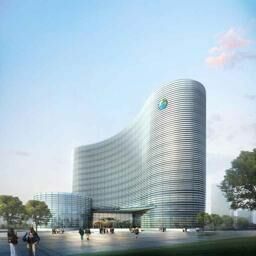}
    \\
    \vspace{-0.5em}
    \makebox[0.092\textwidth]{\footnotesize Input}
    \makebox[0.092\textwidth]{\footnotesize DeepFill v1~\cite{yu2018generative}}
    \makebox[0.092\textwidth]{\footnotesize DeepFill v2~\cite{yu2019free}}
    \makebox[0.092\textwidth]{\footnotesize CTSDG~\cite{guo2021image}}
    \makebox[0.092\textwidth]{\footnotesize LAMA~\cite{wan2021high}}
    \makebox[0.092\textwidth]{\footnotesize MISF~\cite{li2022misf}}
    \makebox[0.092\textwidth]{\footnotesize WNet~\cite{zhang2022w}}
    \makebox[0.092\textwidth]{\footnotesize LDM~\cite{Rombach_2022_CVPR}}
    \makebox[0.092\textwidth]{\footnotesize HINT (Ours)}
    \makebox[0.092\textwidth]{\footnotesize GT}
    \\
    \vspace{0.5em}
    \includegraphics[width=0.092\textwidth]{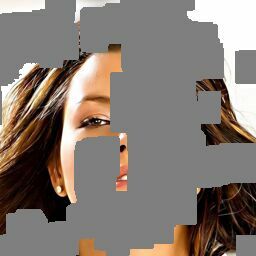}
    \includegraphics[width=0.092\textwidth]{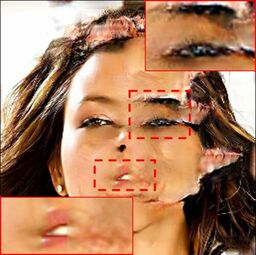}
    \includegraphics[width=0.092\textwidth]{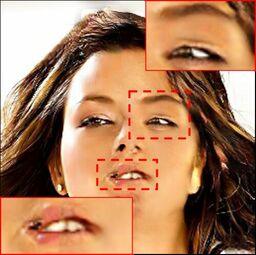}
    \includegraphics[width=0.092\textwidth]{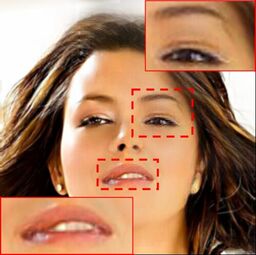}
    \includegraphics[width=0.092\textwidth]{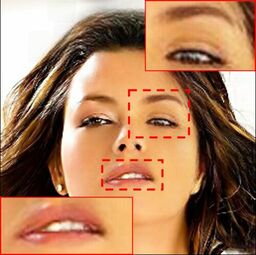}
    \includegraphics[width=0.092\textwidth]{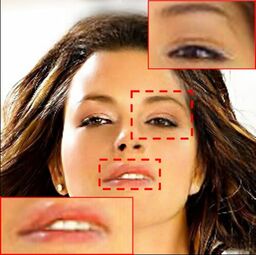}
    \includegraphics[width=0.092\textwidth]{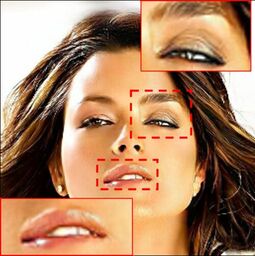}
    \includegraphics[width=0.092\textwidth]{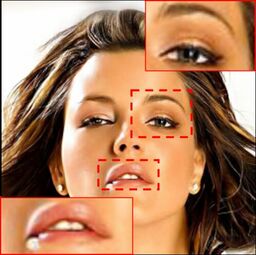}
    \includegraphics[width=0.092\textwidth]{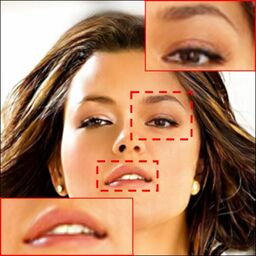}
    \includegraphics[width=0.092\textwidth]{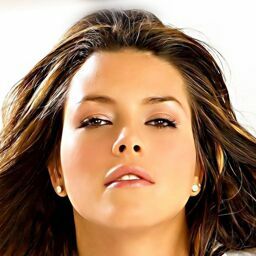}
    \\
    \includegraphics[width=0.092\textwidth]{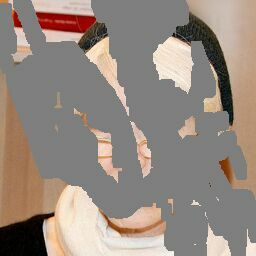}
    \includegraphics[width=0.092\textwidth]{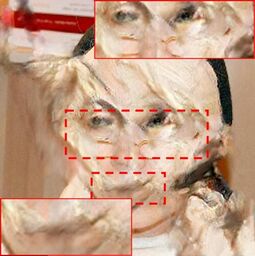}
    \includegraphics[width=0.092\textwidth]{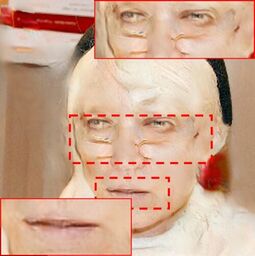}
    \includegraphics[width=0.092\textwidth]{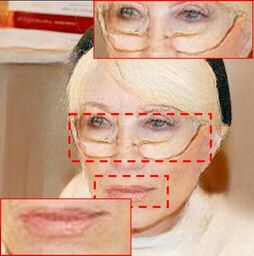}
    \includegraphics[width=0.092\textwidth]{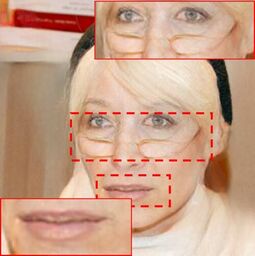}
    \includegraphics[width=0.092\textwidth]{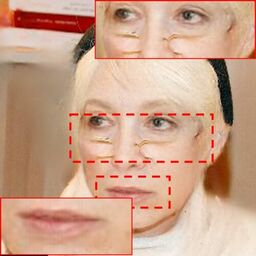}
    \includegraphics[width=0.092\textwidth]{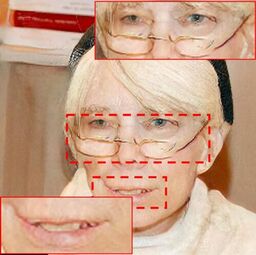}
    \includegraphics[width=0.092\textwidth]{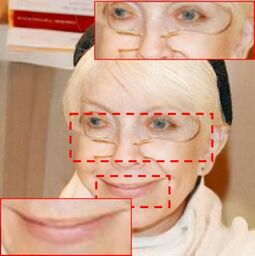}
    \includegraphics[width=0.092\textwidth]{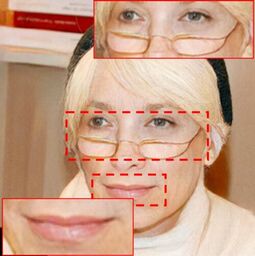}
    \includegraphics[width=0.092\textwidth]{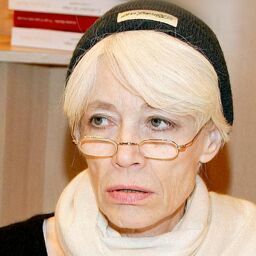}
    \\
    \includegraphics[width=0.092\textwidth]{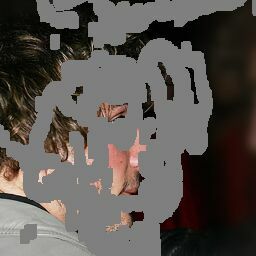}
    \includegraphics[width=0.092\textwidth]{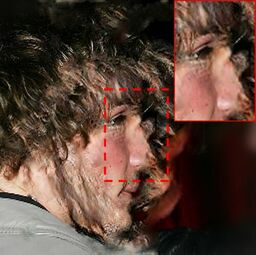}
    \includegraphics[width=0.092\textwidth]{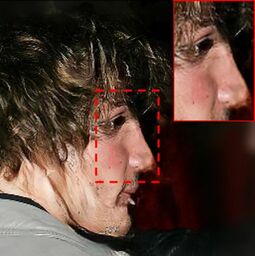}
    \includegraphics[width=0.092\textwidth]{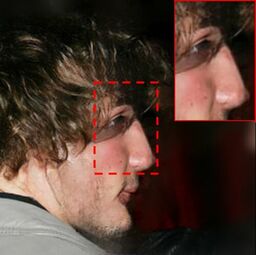}
    \includegraphics[width=0.092\textwidth]{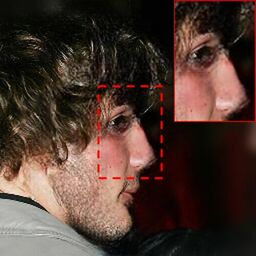}
    \includegraphics[width=0.092\textwidth]{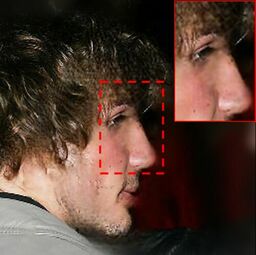}
    \includegraphics[width=0.092\textwidth]{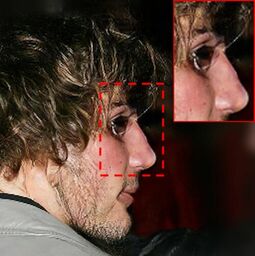}
    \includegraphics[width=0.092\textwidth]{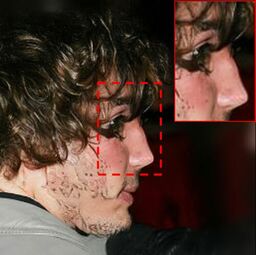}
    \includegraphics[width=0.092\textwidth]{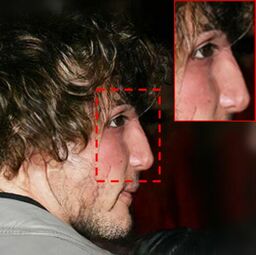}
    \includegraphics[width=0.092\textwidth]{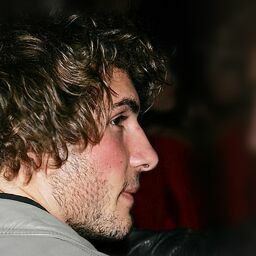}
    \\
    \includegraphics[width=0.092\textwidth]{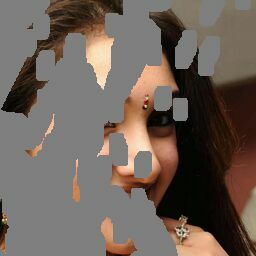}
    \includegraphics[width=0.092\textwidth]{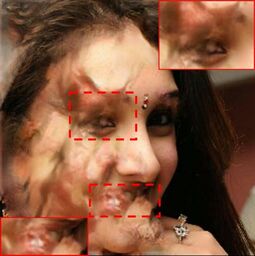}
    \includegraphics[width=0.092\textwidth]{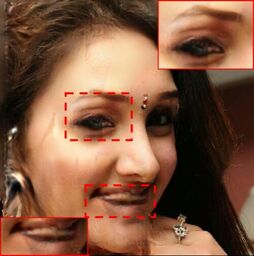}
    \includegraphics[width=0.092\textwidth]{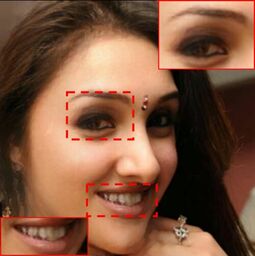}
    \includegraphics[width=0.092\textwidth]{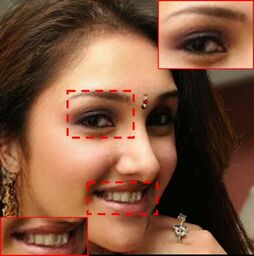}
    \includegraphics[width=0.092\textwidth]{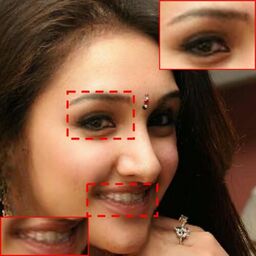}
    \includegraphics[width=0.092\textwidth]{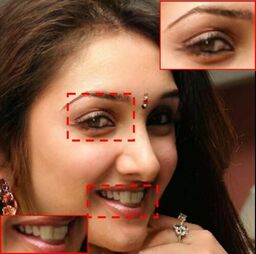}
    \includegraphics[width=0.092\textwidth]{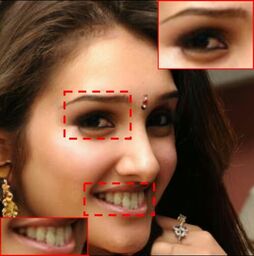}
    \includegraphics[width=0.092\textwidth]{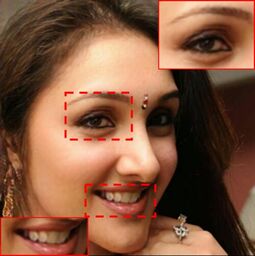}
    \includegraphics[width=0.092\textwidth]{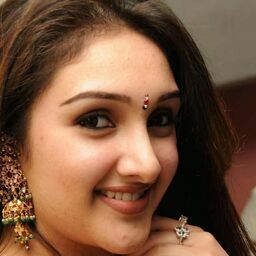}
    \\
    \vspace{-0.5em}
    \makebox[0.092\textwidth]{\footnotesize Input}
    \makebox[0.092\textwidth]{\footnotesize DeepFill v1~\cite{yu2018generative}}
    \makebox[0.092\textwidth]{\footnotesize DeepFill v2~\cite{yu2019free}}
    \makebox[0.092\textwidth]{\footnotesize WaveFill~\cite{yu2021wavefill}}
    \makebox[0.092\textwidth]{\footnotesize LAMA~\cite{wan2021high}}
    \makebox[0.092\textwidth]{\footnotesize WNet~\cite{zhang2022w}}
    \makebox[0.092\textwidth]{\footnotesize MAT~\cite{li2022mat}}
    \makebox[0.092\textwidth]{\footnotesize RePaint~\cite{Lugmayr_2022_CVPR}}
    \makebox[0.092\textwidth]{\footnotesize HINT (Ours)}
    \makebox[0.092\textwidth]{\footnotesize GT}
    \\
    \caption{Comparisons with visualisations $(256 \times 256)$ showing that our results are more coherent in structure and sharper in texture and semantic details. The top two rows are from CelebA-HQ~\cite{karras2017progressive} and the bottom two rows are from Places2~\cite{zhou2017places}.} 
    \label{fig:comparison_CelebAHQ&Places2}
\end{figure*}

\begin{figure}[tb] \centering
    \makebox[0.01\textwidth]{}
    \\
    \raisebox{1.0\height}{\makebox[0.01\textwidth]{\rotatebox{90}{\makecell{\scriptsize Input}}}}
    \includegraphics[width=0.108\textwidth]{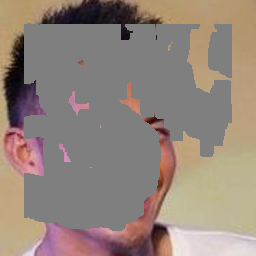}
    \includegraphics[width=0.108\textwidth]{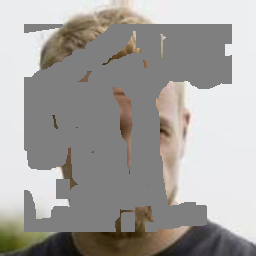}
    \raisebox{1.0\height}{\makebox[0.01\textwidth]{\rotatebox{90}{\makecell{\scriptsize Input}}}}
    \includegraphics[width=0.108\textwidth]{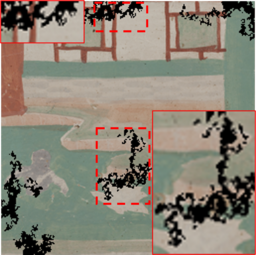}
    \includegraphics[width=0.108\textwidth]{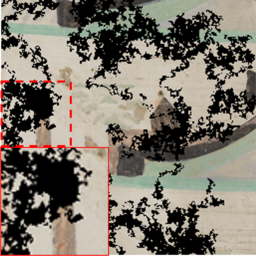}
    \\
    \raisebox{0.2\height}{\makebox[0.01\textwidth]{\rotatebox{90}{\makecell{\scriptsize CTSDG~\cite{guo2021image}}}}}
    \includegraphics[width=0.108\textwidth]{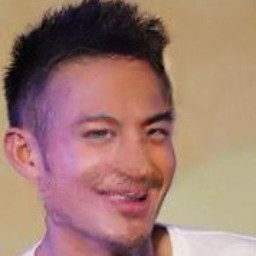}
    \includegraphics[width=0.108\textwidth]{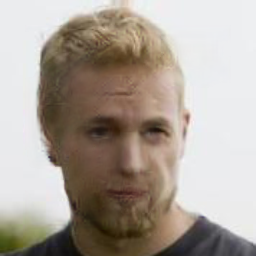}
    \raisebox{0.3\height}{\makebox[0.01\textwidth]{\rotatebox{90}{\makecell{\scriptsize JPGNet~\cite{guo2021jpgnet}}}}}
    \includegraphics[width=0.108\textwidth]{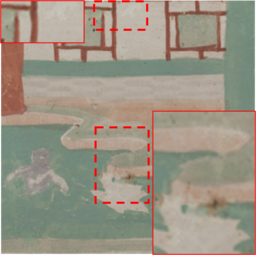}
    \includegraphics[width=0.108\textwidth]{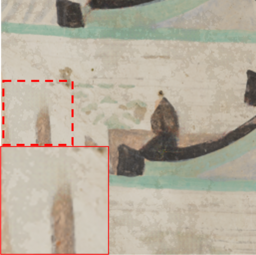}
    \\
    \raisebox{0.3\height}{\makebox[0.01\textwidth]{\rotatebox{90}{\makecell{\scriptsize MISF~\cite{li2022misf}}}}}
    \includegraphics[width=0.108\textwidth]{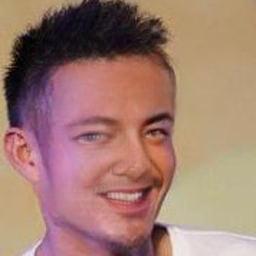}
    \includegraphics[width=0.108\textwidth]{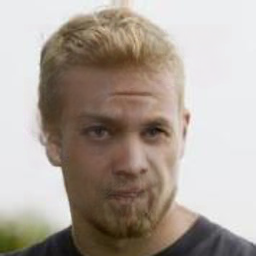}
    \raisebox{0.3\height}{\makebox[0.01\textwidth]{\rotatebox{90}{\makecell{\scriptsize MISF~\cite{li2022misf}}}}}
    \includegraphics[width=0.108\textwidth]{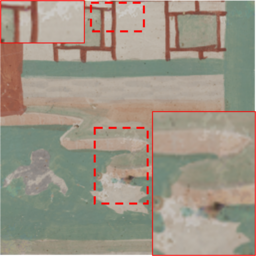}
    \includegraphics[width=0.108\textwidth]{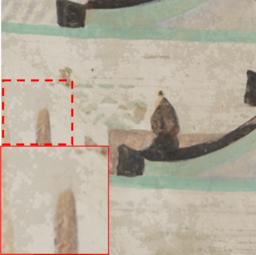}
    \\
    \raisebox{0.8\height}{\makebox[0.01\textwidth]{\rotatebox{90}{\makecell{\scriptsize Ours}}}}
    \includegraphics[width=0.108\textwidth]{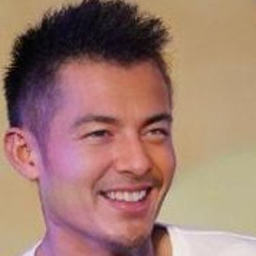}
    \includegraphics[width=0.108\textwidth]{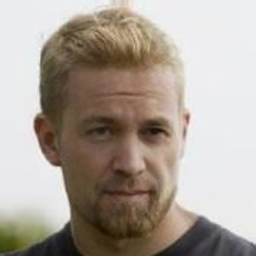}
    \raisebox{0.8\height}{\makebox[0.01\textwidth]{\rotatebox{90}{\makecell{\scriptsize Ours}}}}
    \includegraphics[width=0.108\textwidth]{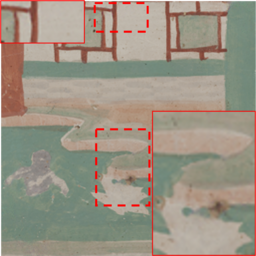}
    \includegraphics[width=0.108\textwidth]{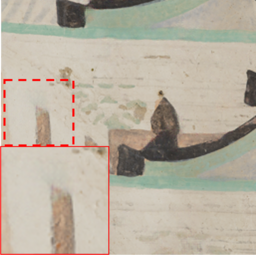}
    \\
    \raisebox{2.5\height}{\makebox[0.01\textwidth]{\rotatebox{90}{\makecell{\scriptsize GT}}}}
    \includegraphics[width=0.108\textwidth]{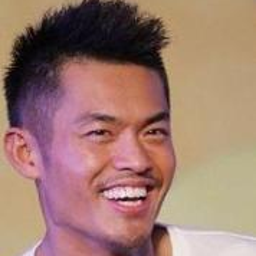}
    \includegraphics[width=0.108\textwidth]{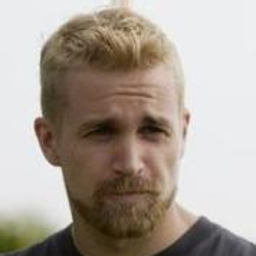}
    \raisebox{2.5\height}{\makebox[0.01\textwidth]{\rotatebox{90}{\makecell{\scriptsize GT}}}}
    \includegraphics[width=0.108\textwidth]{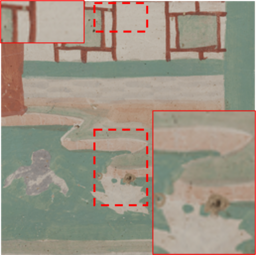}
    \includegraphics[width=0.108\textwidth]{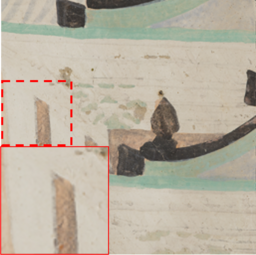}
    \caption{Comparisons with SOTAs~\cite{li2022misf,guo2021jpgnet,guo2021image} on CelebA~\cite{liu2015deep} (masks shown in grey) and Dunhuang~\cite{yu2019dunhuang} (masks shown in black). Red boxes highlight differences.} 
    \vspace{-0.7em}
    \label{fig:compare_CelebA&Dunhuang}
\end{figure}
\newcommand{\PSNRT}{PSNR}
\newcommand{\SSIMT}{SSIM}
\newcommand{\LT}{L1}
\newcommand{\LPIPST}{LPIPS}
\newcommand{\FID}{FID}
\newcommand{\LowExposure}{0.01\%-20\%}
\newcommand{\MiddleExposure}{20\%-40\%}
\newcommand{\HighExposure}{40\%-60\%}
\newcommand{\AllExposure}{Default}

\newcommand{\Frst}[1]{{\textbf{#1}}}
\newcommand{\Scnd}[1]{{\underline{#1}}}

\begin{table*}[t] 
\centering
\captionsetup[table]{skip=4pt}
\caption{Comparison results on (a, top) CelebA-HQ, (b, middle) CelebA and (c, bottom) Places2. The \textbf{bold} and \underline{underline} indicate the best and the second best respectively.}
\resizebox{\textwidth}{!}{
    {
    \begin{tabular}{@{\extracolsep{3pt}}c c c c c c| c c c c c| c c c c c@{}}
        \hline\hline
        \textbf{CelebA-HQ}
        & \multicolumn{5}{c}{\LowExposure} & \multicolumn{5}{c}{\MiddleExposure} & \multicolumn{5}{c}{\HighExposure} \T\\
        Method &\PSNRT$\uparrow$ & \SSIMT$\uparrow$ & \LT$\downarrow$ & \FID$\downarrow$ & \LPIPST$\downarrow$ & \PSNRT$\uparrow$ & \SSIMT$\uparrow$ & \LT$\downarrow$ & \FID$\downarrow$ & \LPIPST$\downarrow$ & \PSNRT$\uparrow$ & \SSIMT$\uparrow$ & \LT$\downarrow$ & \FID$\downarrow$ & \LPIPST$\downarrow$  \\
        \hline

        DeepFill v1 \cite{yu2018generative} & 34.2507 & 0.9047 & 1.7433 & 2.2141 & 0.1184 & 26.8796 & 0.8271 & 2.3117 & 9.4047 & 0.1329 & 21.4721 & 0.7492 & 4.6285 & 15.4731 & 0.2521\T\\

        DeepFill v2 \cite{yu2019free} & 34.4735 & 0.9533 & 0.5211 & 1.4374 & 0.0429 & 27.3298 & 0.8657 & 1.7687 & 5.5498 & 0.1064 & 22.6937 & 0.7962 & 3.2721 & 8.8673 & 0.1739\\

        LaMa \cite{suvorov2022resolution} & \Scnd{35.5656} & 0.9685 & 0.4029 & 1.4309 & 0.0319 & \Scnd{28.0348} & 0.8983 & \Scnd{1.3722} & 4.4295 & 0.0903 & \Scnd{23.9419} & \Scnd{0.8003} & \Scnd{2.8646} & 8.4538 & 0.1620\\

        WNet \cite{zhang2022w} & 35.3591 & 0.9647 & 0.4957 & 1.2759 & 0.0287 & 28.1736 & 0.8872 & 1.4495 & 4.7299 & 0.0833 & 23.8357 & 0.7872 & 2.9316 & 9.4926 & 0.1649\\
        
        MAT \cite{li2022mat}  & 35.5466 & \Scnd{0.9689} & \Scnd{0.3961} & \Scnd{1.2428} & \Scnd{0.0268} & 27.6684 & \Scnd{0.8957} & 1.3852 & \Scnd{3.4677} & \Scnd{0.0832} & 23.3371 & 0.7964 & 2.9816 & \Scnd{5.7284} & \Scnd{0.1575}\\

        WaveFill \cite{yu2021wavefill} & 31.4695 & 0.9290 & 1.3228 & 6.0638 & 0.0802 & 27.1073 & 0.8668 & 2.1159 & 8.3804 & 0.1231 & 23.3569 & 0.7817 & 3.5617 & 13.0849 & 0.1917\B\\
        \hline
        Ours & \Frst{36.5725} & \Frst{0.9777} & \Frst{0.3942} & \Frst{1.1128} & \Frst{0.0228} & \Frst{28.6247} & \Frst{0.9195} & \Frst{1.2885} &\Frst{3.3915} & \Frst{0.0745} & \Frst{24.1287} & \Frst{0.8241} & \Frst{2.7778} & \Frst{5.6179} & \Frst{0.1449}\B\T\\
        \hline\hline


        \textbf{Places2} & \multicolumn{5}{c}{\LowExposure} & \multicolumn{5}{c}{\MiddleExposure} & \multicolumn{5}{c}{\HighExposure} \T\\
         Method & \PSNRT$\uparrow$ & \SSIMT$\uparrow$ & \LT$\downarrow$ &\FID$\downarrow$ & \LPIPST$\downarrow$ & \PSNRT$\uparrow$ & \SSIMT$\uparrow$ & \LT$\downarrow$ &\FID$\downarrow$ & \LPIPST$\downarrow$ & \PSNRT$\uparrow$ & \SSIMT$\uparrow$ & \LT$\downarrow$ &\FID$\downarrow$ & \LPIPST$\downarrow$   \\
        \hline

        DeepFill v1\cite{yu2018generative} & 30.2958 & 0.9532 & 0.6953 & 26.3275 & 0.0497 & 24.2983 & 0.8426 & 2.4927 & 31.4296 & 0.1472 & 19.3751 & 0.6473 & 5.2092 & 46.4936 & 0.3145\T\\

        DeepFill v2\cite{yu2019free} & 31.4725 & 0.9558 & 0.6632 & 23.6854 & 0.0446 & 24.7247 & 0.8572 & 2.2453 & 27.3259 & 0.1362 & 19.7563 & 0.6742 & 4.9284 & 36.5458 & 0.2891\\

        CTSDG \cite{guo2021image}  & 32.111 & 0.9565 & 0.6216 & 24.9852 & 0.0458 & 24.6502 & 0.8536 & 2.1210 &  29.2158 & 0.1429 & 20.2962 & 0.7012 & 4.6870 & 37.4251 & 0.2712\\

        WNet \cite{zhang2022w} & 32.3276 & 0.9372 & 0.5913 & 20.4925 & 0.0387 & 25.2198 & 0.8617 & 2.0765 & 24.7436 & 0.1136 & 20.4375 & 0.6727 & 4.6371 & 32.6729 & 0.2416\\

        MISF~ \cite{li2022misf}  & \Scnd{32.9873} & \Scnd{0.9615} & \Scnd{0.5931} & 21.7526 & 0.0357 & \Scnd{25.3843} & \Scnd{0.8681} & \Scnd{1.9460} & 30.5499  & 0.1183 & \Scnd{20.7260} & 0.7187 & 4.4383 & 44.4778 & 0.2278 \\

        LaMa \cite{suvorov2022resolution}  & 32.4660 & 0.9584 & 0.5969 & \Scnd{14.7288} & \Scnd{0.0354} & 25.0921 & 0.8635 & 2.0048 & \Scnd{22.9381} & \Scnd{0.1079} & 20.6796 & \Scnd{0.7245} & \Scnd{4.4060} & \Scnd{25.9436} & \Scnd{0.2124}\\   

        WaveFill \cite{yu2021wavefill} & 29.8598 & 0.9468 & 0.9008 &30.4259 & 0.0519 & 23.9875 & 0.8395 & 2.5329 & 39.8519 & 0.1365 & 18.4017 & 0.6130 & 7.1015 & 56.7527 & 0.3395\B\\
        \hline
        Ours    & \Frst{33.0276} & \Frst{0.9689} & \Frst{0.5612} & \Frst{13.9128} & \Frst{0.0307} & \Frst{25.4216} & \Frst{0.8807} & \Frst{1.9270} & \Frst{20.0241} & \Frst{0.1003} & \Frst{20.9243} & \Frst{0.7470} & \Frst{4.3296} & \Frst{25.7150} & \Frst{0.2041} \T\B\\
        \hline
        \hline
        \textbf{CelebA} & \multicolumn{5}{c}{\LowExposure} & \multicolumn{5}{c}{\MiddleExposure} & \multicolumn{5}{c}{\HighExposure} \T\\
        Method &\PSNRT$\uparrow$ & \SSIMT$\uparrow$ & \LT$\downarrow$ &\FID$\downarrow$ & \LPIPST$\downarrow$ & \PSNRT$\uparrow$ & \SSIMT$\uparrow$ & \LT$\downarrow$ &\FID$\downarrow$ & \LPIPST$\downarrow$ & \PSNRT$\uparrow$ & \SSIMT$\uparrow$ & \LT$\downarrow$ & \FID$\downarrow$ & \LPIPST$\downarrow$  \\
        \hline

        CTSDG \cite{guo2021image}  & 36.465 & 0.9732 & 0.5871 & \Scnd{2.5876} & 0.0334& 29.1393 & \Scnd{0.9159} & 1.38 & \Scnd{7.4925} & 0.0935 & 23.8371 & \Scnd{0.8157} & \Scnd{3.04} &  \Scnd{9.8473} & \Scnd{0.1815}\T\\

        MISF \cite{li2022misf}      & \Scnd{36.8981} & \Scnd{0.9747} & \Scnd{0.3441} & 3.3598& \Scnd{0.0333} & \Scnd{28.9270} & 0.9103 & \Scnd{1.227} & 8.0249 & \Scnd{0.1031} & \Scnd{23.5355} & 0.8033 & 3.182 & 13.2475 & 0.2012\B\\ 
        \hline

        Ours & \Frst{37.5696} & \Frst{0.9754} &\Frst{0.3402} & \Frst{1.0270} & \Frst{0.0232} & \Frst{29.8525} & \Frst{0.9208} & \Frst{1.220} & \Frst{4.1359}& \Frst{0.0689} & \Frst{24.4538} & \Frst{0.8270} & \Frst{2.7802} & \Frst{5.3612} & \Frst{0.1408}\T\B\\
        \hline\hline
    \end{tabular}
    }
    }
    \label{tab:res_real_data_wide}
    \label{tab:compare}
\end{table*}
\begin{table}\centering
\captionsetup[table]{skip=4pt}
\caption{Comparisons on the Dunhuang Challenge dataset.} 
\scriptsize
\resizebox{0.40\textwidth}{!}{
    {
    \tabulinesep=0mm
    \begin{tabular}{*{5}{c}}
    \hline\hline
    Model &  PSNR$\uparrow$ & SSIM$\uparrow$ &  L1$\downarrow$ & LPIPS$\downarrow$\T\B\\
    \hline
    StructFlow~\cite{ren2019structureflow} & 35.199& 0.9559 & 0.475 & 0.0589\T\\
    EdgeConnect~\cite{nazeri2019edgeconnect} & 36.419 & 0.9635 & 0.441 & 0.0480 \\
    RFRNet~\cite{li2020recurrent} & 36.485 & 0.9648 & 0.401 & 0.0463 \\
    JPGNet~\cite{guo2021jpgnet} &37.646 & 0.9724 & 0.353 & 0.0469\\
    MISF~\cite{li2022misf} & 38.383 & 0.9735 & 0.341 & 0.0330\B\\
    \hline
    Ours & \textbf{38.6705} & \textbf{0.9743} & \textbf{0.3161} & \textbf{0.0286}\T\B\\
    \hline\hline
    \end{tabular}
    }
    }
    \label{tab:dunhuang}
\end{table}

\begin{table}
\begin{center}
\caption{Number of parameter and inference time}
\addtolength{\tabcolsep}{-2pt}
\resizebox{0.60\columnwidth}{!}{
\begin{tabular}{ccc}
\hline\hline
Model & Param $\times 10^6$ & Infer. Time/per img\T\B\\
\hline
DeepFill v1 \cite{yu2018generative} & 3 & 7 $ms$\T \\
DeepFill v2 \cite{yu2019free} & 4 & 10 $ms$ \\
Wavefill \cite{yu2021wavefill} & 49 & 70 $ms$ \\
CTSDG \cite{guo2021image} & 52 & 20 $ms$\\
WNet \cite{zhang2022w} & 46 & 35 $ms$\\
MISF \cite{li2022misf} & 26 & 10 $ms$ \\
MAT \cite{li2022mat} & 62 & 70 $ms$ \\
LAMA \cite{suvorov2022resolution} & 51 & 25 $ms$ \\
Stable Diffusion & 860 & 880 $ms$\\
LDM~\cite{Rombach_2022_CVPR} & 387 & ~6000 $ms$\\
Repaint~\cite{Lugmayr_2022_CVPR} & 552 & ~250000 $ms$\B\\
\hline
{Ours} & {139} & {125} $ms$\T\B\\
\hline\hline
\end{tabular}
}
\addtolength{\tabcolsep}{1pt}
\vspace{-0.5em}
\label{tab:param}
\end{center}
\end{table}

\section{Experiments}
In this section, we present a comprehensive evaluation of the proposed HINT. First, we describe the datasets employed and delve into the specifics of the implementation. Then, we compare HINT with state-of-the-art methods to showcase its superior performance, with both quantitative and qualitative results. Finally, we conduct thorough ablation studies to evaluate the significance of each proposed component.

\subsection{Datasets}
To assess the efficacy of our proposed method, we employ CelebA \cite{liu2015deep}, CelebA-HQ \cite{karras2017progressive}, Places2-Standard \cite{zhou2017places} and Dunhuang Challenge \cite{yu2019dunhuang} datasets. All experiments are conducted with 256$\times$256 images, providing a comprehensive evaluation of our approach in a consistent and well-defined setting.
The CelebA~\cite{liu2015deep} and CelebA-HQ~\cite{karras2017progressive} are two human face datasets with different qualities, while the Places2-Standard dataset is a subset of the Places2~\cite{zhou2017places} dataset offering a diverse collection of scenes, such as indoor and outdoor environments, natural landscapes, and man-made structures and constructions. These three datasets are commonly used within the existing literature on inpainting~\cite{zheng2022bridging, wan2021high, li2022mat}, making them ideal for evaluating our approach. The Dunhuang Challenge~\cite{yu2019dunhuang} dataset represents a practical application of image inpainting in real-world scenarios. 

For CelebA and Dunhuang, we follow the standard configuration to split the data for training and testing. In the case of the CelebA-HQ dataset, to ensure reproducibility, we use the first 28,000 images for training and the remaining 2,000 images for testing. For the Places2-Standard dataset, we use the standard training set and validation set for training and testing, respectively. For mask settings, we follow prior work~\cite{guo2021image, li2022misf} and use irregular masks~\cite{liu2018image} for CelebA, CelebA-HQ, and Places2. As for Dunhuang Challenge, we use the officially released masks for testing.

\subsection{Implementation Details}
In the 7-level transformer blocks, the number of Sandwich blocks is sequentially set to [4,6,6,8,6,6,4] and the attention head in SCAL are [1,2,4,8,4,2,1]. \revision{All experiments are carried out on a single NVidia A100 GPU with a batch size of 4. We adopt the Adam optimiser~\cite{kingma2020method} with $\beta_1=0.9$, $\beta_2=0.999$. The learning rate is initially set to $1 e^{-4}$ and is halved at the 75\% milestone of the training progress. Compared to the state of the art in the existing literature \cite{guo2021image,wan2021high,zeng2019learning}, our approach is more robust against small changes in the training procedure, making it more generalisable and easier to deploy. Our training pipeline does not rely on warm-up step \cite{wan2021high}, pre-training requirements \cite{zeng2019learning} or fine-tuning \cite{guo2021image}.}

\subsection{Comparison with the State of the Art}


In assessing our HINT, designed to generate high-quality, fine-grained images, we follow \cite{li2022misf} to employ a suite of evaluation metrics: Peak Signal-to-Noise Ratio (PSNR), Structural Similarity (SSIM), L1 and Perceptual Similarity (LPIPS). These chosen metrics align with our intent to create a nuanced and comprehensive understanding of the performance of models. PSNR and L1 are used to measure pixel-wise reconstruction accuracy, which reflects the fidelity of the inpainted output. SSIM~\cite{wang2004image} evaluates structural similarity, ensuring the inpainted segments remain coherent within the image contextually. We also include LPIPS~\cite{zhang2018unreasonable}, a learned perceptual metric, capable of detecting complex distortions that mirror human perceptual differences, a crucial attribute when the aim is to produce high-quality imagery.

We categorise the masks into three groups based on the mask ratio, i.e., small (0.01\%-20\%), medium (20\%-40\%) and large (40\%-60\%), referring to the extent of missing regions.

\textbf{Quantitative Results} As shown in Tab.~\ref{tab:compare} and Tab.~\ref{tab:dunhuang}, HINT achieves a better overall performance across all datasets and mask ratios than the state of the arts
\cite{guo2021jpgnet,li2022misf,guo2021image,wan2021high,yu2021wavefill,li2022mat}. Compare to the latest transformer-based MAT~\cite{li2022mat} on CelebA-HQ, HINT improves PSNR by 5.7\%, 3.3\% and 3.4\% at the increasing mask ratios respectively, demonstrating that it preserves more high-fidelity details in reconstructed images. In Places2, compared with the latest high-quality inpainting method MISF~\cite{li2022misf}, HINT achieves a 12.6\%, 13.8\% and 7.2\% decrease for LPIPS, showcasing its effectiveness in perceptual recovery. Since the Dunhuang Challenge provides standard masks, we crawled the benchmark from \cite{li2022misf} for comparison. HINT outperforms existing models across all metrics.

For a comprehensive and robust evaluation, we also compare our model with the state-of-the-art diffusion model-based methods with large masks, which are well-known for their prowess in generating high-quality images~\cite{chang23design}. \secrevision{Three prominent diffusion models, LDM \cite{Rombach_2022_CVPR}, Stable Diffusion (SD) and RePaint \cite{Lugmayr_2022_CVPR}, are chosen for comparison. To allow for a fair comparison, all experiments are conducted on officially released pretrained models on the corresponding datasets. It is important to note that SD does not provide models pretrained on either CelebA-HQ or Places2, so, we chose the LAION v2 5+ pre-trained model, as its data distribution is similar to that of the Places2 dataset, but it is much larger and of higher quality.  Tab.~\ref{tab:diffusion} and Tab.~\ref{tab:param} underscore the superior performance of our model across all metrics and signify the efficiency in image inpainting tasks.}
Ideally, we wish to assess all diffusion model on Places2. However, due to the significant inference time required by RePaint (Tab.~\ref{tab:param}), a single evaluation on the Places2 dataset for three mask ratios demands around one GPU-year, making it computationally intractable. As a result, we chose to evaluate LDM on Places2, given its relatively more manageable inference time, and focused our analysis of RePaint on the CelebA-HQ dataset.





\begin{table}\centering
\captionsetup[table]{skip=4pt}
\caption{Comparison with diffusion models.}
\label{tab:diffusion}
\scriptsize
\resizebox{0.40\textwidth}{!}{
    {
    
    \tabulinesep=0mm
    \begin{tabular}{*{6}{c}}
    \hline\hline
    Plces2 &  PSNR$\uparrow$ & SSIM$\uparrow$ &  L1$\downarrow$ & FID$\downarrow$ & LPIPS$\downarrow$\T\B\\
    \hline
    LDM~\cite{Rombach_2022_CVPR} & 19.6476 & 0.7052 & 4.6895 & 27.3619 & 0.2675\T\\
    Stable Diffusion$^*$ & 19.4812 & 0.7185 & 4.5729 & 27.8830 & 0.2416\\
    Ours & \textbf{20.8579} & \textbf{0.7227} & \textbf{4.3814} & \textbf{26.7895} & \textbf{0.2102} \\
    \hline
    \hline
    CelebA-HQ &  PSNR$\uparrow$ & SSIM$\uparrow$ &  L1$\downarrow$ & FID$\downarrow$ & LPIPS$\downarrow$\T\B\\
    \hline
    RePaint~\cite{Lugmayr_2022_CVPR} & 21.8321 & 0.7791 & 3.9427 & 8.9637 & 0.1943 \T\\
    Ours &\textbf{24.1287} & \textbf{0.8241} & \textbf{2.778} & \textbf{7.5793} &\textbf{0.1449}\\
    \hline\hline
    \multicolumn{6}{l}{\small $^*$: The officially released Stable Diffusion inpainting model} \T\\
    \multicolumn{6}{l}{\small pretrained on high-quality LAION-Aesthetics V2 5+ dataset.} \\
    \end{tabular}
    \vspace{-4.5em}
    }
    }
\end{table}
\textbf{Qualitative Results}
We provide the exemplar visual results to further demonstrate the advantages of HINT over comparators. As shown in Fig.~\ref{fig:comparison_CelebAHQ&Places2}, our model generates high-quality images with more coherent structures and fewer artifacts, such as roofs and planks. For face restoration, our model better recovers finer-grained details, such as eye features, compared to the current state of the art~\cite{wan2021high,yu2021wavefill,li2022mat}.
We also provide qualitative results for CelebA~\cite{liu2015deep} and Dunhuang datasets~\cite{yu2019dunhuang} in Fig.~\ref{fig:compare_CelebA&Dunhuang} to indicate our superior performance in global context modeling. The proposed HINT recovers high-quality faces with clear textures and plausible semantics, even with a large mask covering almost all facial attributes. The results on Dunhuang show that our model suppresses the generation of light mottle, and demonstrates the effectiveness of our model in handling small scratch masks.

\textbf{Efficiency Comparison}
Our model uniquely incorporates spatial awareness into the channel-wise self-attention, a design innovation that maintains linear complexity, $\mathcal{O}(C^2)$, with $C$ being the channel number. It manages to strike an impressive balance between complexity and efficiency. As shown in Tab.~\ref{tab:param}, our model, carrying 139 million parameters, still situates itself within the parameter counts seen among state-of-the-art methods. More significantly, our model upholds an inference time of 125ms per image, ensuring practicality with millisecond-level response time. This efficiency does not come at the expense of performance since our model outshines competing methods in both qualitative and quantitative evaluations.
\begin{figure}[t] \centering
    \includegraphics[width=0.06\textwidth]{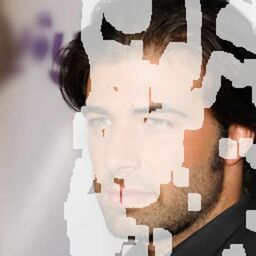}
    \includegraphics[width=0.06\textwidth]{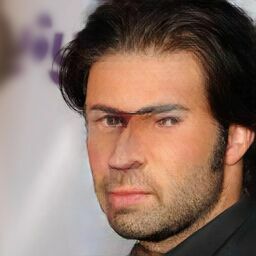}
    \includegraphics[width=0.06\textwidth]{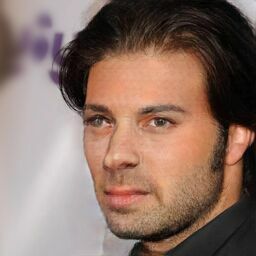}
    \includegraphics[width=0.06\textwidth]{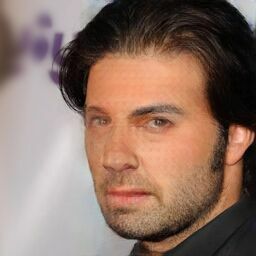}
    \includegraphics[width=0.06\textwidth]{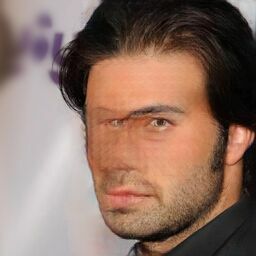}
    \includegraphics[width=0.06\textwidth]{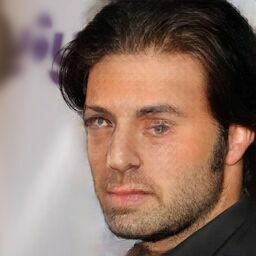}
    \includegraphics[width=0.06\textwidth]{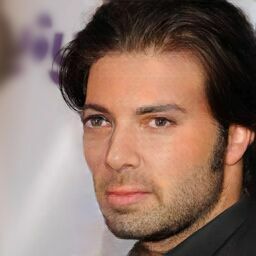}
    \\
    \makebox[0.06\textwidth]{\footnotesize Masked Input}
    \makebox[0.06\textwidth]{\footnotesize A}
    \makebox[0.06\textwidth]{\footnotesize B}
    \makebox[0.06\textwidth]{\footnotesize C}
    \makebox[0.06\textwidth]{\footnotesize D}
    \makebox[0.06\textwidth]{\footnotesize E}
    \makebox[0.06\textwidth]{\footnotesize  Ours}
    \\
    \caption{{Visual results of our ablation studies. A refers to replacing MPD with conventional PD, B removes the first FFN in \enquote{sandwich}, C replaces SCAL with a single channel-wise self-attention design, D ablates HINT to only include channel self-attention, a single FFN, and convolutional down-sampling. E replaces our spatial branch with the basic gated mechanism from~\cite{deng2022t}.}} 
    \vspace{-0.5em}
    \label{fig:ablation}
\end{figure}
\begin{table*}[t]
\captionsetup[table]{skip=4pt}
\caption{Ablation studies. Setup A replaces MPD with conventional PD, B removes the first FFN in \enquote{Sandwich}, C replaces SCAL with single channel-wise self-attention design, D is a HINT variant with the spatial branch replaced by \cite{deng2022t}'s gated mechanism.}
	\centering
		\resizebox{2\columnwidth}{!}{
		{\tabulinesep=0mm
			\begin{tabu}{@{\extracolsep{3pt}} c c c c c c c | c c c c c | c c c c c@{}}
				\hline\hline
				\multicolumn{1}{c}{\multirow{2}{*}{Setup}} & \multicolumn{1}{c}{\multirow{2}{*}{Model}} & 
				\multicolumn{5}{c}{0.01\%-20\%} &
				\multicolumn{5}{c}{20\%-40\%} &
				\multicolumn{5}{c}{40\%-60\%}\T\B \\
				\cline{3-7} \cline{8-12} \cline{13-17}
						&& PSNR$\uparrow$ & SSIM$\uparrow$ & L1$\downarrow$ & FID$\downarrow$ & LPIPS$\downarrow$ & PSNR$\uparrow$ & SSIM$\uparrow$ & L1$\downarrow$ & FID$\downarrow$ & LPIPS$\downarrow$ & PSNR$\uparrow$ & SSIM$\uparrow$ & L1$\downarrow$ & FID$\downarrow$  & LPIPS$\downarrow$\T\B\\
				\hline\hline

        A& w/o MPD   & 34.5955 & 0.9649 & 0.4780 & 1.7458 & 0.0381 & 26.9292 & 0.8863 & 1.6320 & 4.9815 & 0.1084 & 22.5618 & 0.7813 & 3.4982 & 8.2196 & 0.1951\T\\
        B& w/o Sandwich  & 34.7272 & 0.9658 & 0.4661 & 1.4687 & 0.0361 & 27.0914 & 0.8893 & 1.5796 & 4.7625 & 0.1050 & 22.7027 & 0.7853 & 3.4185 & 7.9138 & 0.1912 \\
        C& w/o SCAL & 34.7951 & 0.9659 & 0.4624 & 1.7568 & 0.0364 & 27.1193 & 0.8895 & 1.5732 & 4.8769 & 0.1057 & 22.7206 & 0.7856 & 3.4021 & 8.1627 & 0.1925 \\ 
        D & U-Net w self-attention & 34.0204 & 0.9538 & 0.5129 & 2.0152 & 0.0497 & 26.0814 & 0.8754 & 1.8547 & 5.1029 & 0.1277 & 21.6149 & 0.7679 & 3.6912 & 8.9314 & 0.2104\B\\
        \hline
        E & Full$^{\dag}$  & \text{34.3155} & \text{0.9636}&	\text{0.4891} &	\text{1.3968} &	\text{0.0393} &	\text{26.7534} &	\text{0.8837} &	\text{1.6521} &	\text{4.7358} &	\text{0.1122} &	\text{22.4632} & \text{0.7772} & \text{3.5221} & \text{7.9637} & \text{0.1999} \B\T\\
        \hline
        Ours & Full  & \textbf{35.0436} & \textbf{0.9671}&	\textbf{0.4489} &	\textbf{1.3542} &	\textbf{0.0345} &	\textbf{27.2954} &	\textbf{0.8924} &	\textbf{1.5363} &	\textbf{4.6891} &	\textbf{0.1016} &	\textbf{22.8473} & \textbf{0.7895} & \textbf{3.3403} & \textbf{7.8697} & \textbf{0.1867} \B\T\\

\hline
\hline
\end{tabu}
}
}
\vspace{-0.5em}
\label{tab:ablation}
\end{table*}

\begin{table*}[t]\centering
\captionsetup[table]{skip=4pt}
    \caption{\enquote{Attention-FFN} structure vs. \enquote{FFN-Attention-FFN} structure (Sandwich) with the same number of parameters.} %
    \resizebox{0.99\textwidth}{!}{
    {
    \tabulinesep=0mm
    \begin{tabular}{@{\extracolsep{3pt}}c c c c c c| c c c c c| c c c c c@{}}
        \hline\hline
        \multirow{2}{*}{Model} & \multicolumn{5}{c}{0.01\%-20\%} & \multicolumn{5}{c}{20\%-40\%} & \multicolumn{5}{c}{40\%-60\%} \T\B\\
        \cline{2-6} \cline{7-11} \cline{12-16}
          & PSNR$\uparrow$ & SSIM$\uparrow$ & L1$\downarrow$ & FID$\downarrow$ & LPIPS$\downarrow$ & PSNR$\uparrow$ & SSIM$\uparrow$ & L1$\downarrow$ & FID$\downarrow$ & LPIPS$\downarrow$ & PSNR$\uparrow$ & SSIM$\uparrow$ & L1$\downarrow$ & FID$\downarrow$ & LPIPS$\downarrow$  \T\B\\
        \hline
        SCAL-FFN  & 34.7272 & 0.9658 & 0.4661 & 1.3716 & 0.0361 & 27.0914 & 0.8893 & 1.5796 & 4.7174 & 0.1050 & 22.7027 & 0.7853 & 3.4185 & 7.8970 & 0.1912 \T\\  
        Conformer & 34.5125 & 0.9576 & 0.4729 & 1.4028 & 0.3914 & 26.9672 & 0.8804 & 1.6760 & 4.7597 & 0.1083 & 21.2186 &	0.7218 & 3.6829 & 8.9506 &0.2147\\
        Thin-Sandwich (Ours)  & \textbf{34.7843} & \textbf{0.9661}&	\textbf{0.4614} &	\textbf{1.3697} &	\textbf{0.0357} &	\textbf{27.1070} &	\textbf{0.8911} &	\textbf{1.5763} &	\textbf{4.6993} &	\textbf{0.1047} &	\textbf{22.7075} & \textbf{0.7872} & \textbf{3.4077} & \textbf{7.8863} & \textbf{0.1908} \B\\
        \hline\hline
    \end{tabular}
    }
    }
    \vspace{-0.5em}
    \label{tab:thin_sandwich}
\end{table*}

\begin{table*}[t]\centering
\captionsetup[table]{skip=0pt}
 \caption{Ablation study of using $1\times1$ convolution after the last skip connection.} %
    \resizebox{0.99\textwidth}{!}{
    {
    \tabulinesep=0mm
    \begin{tabular}{@{\extracolsep{3pt}}c c c c c c |c c c c c |c c c c c@{}}
        \hline\hline
        \multirow{2}{*}{Model} & \multicolumn{5}{c}{0.01\%-20\%} & \multicolumn{5}{c}{20\%-40\%} & \multicolumn{5}{c}{40\%-60\%} \T\B\\
        \cline{2-6} \cline{7-11} \cline{12-16}
          & PSNR$\uparrow$ & SSIM$\uparrow$ & L1$\downarrow$ & FID$\downarrow$ & LPIPS$\downarrow$ & PSNR$\uparrow$ & SSIM$\uparrow$ & L1$\downarrow$ & FID$\downarrow$ & LPIPS$\downarrow$ & PSNR$\uparrow$ & SSIM$\uparrow$ & L1$\downarrow$ & FID$\downarrow$ & LPIPS$\downarrow$  \T\B\\
        \hline
        w $1\times1$ conv  & 34.5246 & 0.9646 & 0.4780 & 1.3693  & 0.0386 & 26.8984 & 0.8863 & 1.6267 & 4.7131 & 0.1101 & 22.4694 & 0.7792 & 3.5434 & 7.9647 & 0.1997 \T\\  
        w/o $1\times1$ conv (Ours)  & \textbf{35.0436} & \textbf{0.9671}&	\textbf{0.4489} &	\textbf{1.3542} &	\textbf{0.0345} &	\textbf{27.2954} &	\textbf{0.8924} &	\textbf{1.5363} &	\textbf{4.6891}&	\textbf{0.1016} &	\textbf{22.8473} & \textbf{0.7895} & \textbf{3.3403} & \textbf{7.8697} & \textbf{0.1867} \B\\
        \hline\hline
    \end{tabular}
    }
    }
    \vspace{-0.5em}
    \label{tab:last-skip-connect}
\end{table*} 
\begin{table*}[t]
\captionsetup[table]{skip=4pt}
\centering
\caption{Different kernal size in the embedding layer.}
\resizebox{2\columnwidth}{!}{
{
    \tabulinesep=0mm
    \begin{tabu}{@{\extracolsep{3pt}}c c c c c c |c c c c c |c c c c c@{}}
    \hline\hline
    \multicolumn{1}{c}{\multirow{2}{*}{Model}} & 
    \multicolumn{5}{c}{0.01\%-20\%} &
    \multicolumn{5}{c}{20\%-40\%} &
    \multicolumn{5}{c}{40\%-60\%}\T\B \\
    \cline{2-6} \cline{7-11} \cline{11-16}
         & PSNR$\uparrow$ & SSIM$\uparrow$ & L1$\downarrow$ & FID$\downarrow$ & LPIPS$\downarrow$ & PSNR$\uparrow$ & SSIM$\uparrow$ & L1$\downarrow$ & FID$\downarrow$ & LPIPS$\downarrow$ & PSNR$\uparrow$ & SSIM$\uparrow$ & L1$\downarrow$ & FID$\downarrow$ & LPIPS$\downarrow$\T\B\\
        \hline\hline
        $7\times7$ emb   & 34.6389 & 0.9657 & 0.4681 & 1.4034 & 0.0366 & 27.0422 & 0.8905 & 1.5803 & 4.9783 & 0.1043 & 22.6667 & 0.7865 & 3.3950 & 7.9168 & 0.1898\T\\
        $3\times3$ emb (Ours)  & \textbf{35.0436} & \textbf{0.9671}&	\textbf{0.4489} &	\textbf{1.3542} &	\textbf{0.0345} &	\textbf{27.2954} &	\textbf{0.8924} &	\textbf{1.5363} &	\textbf{4.6491} &	\textbf{0.1016} &	\textbf{22.8473} & \textbf{0.7895} & \textbf{3.3403} &	\textbf{7.8697} & \textbf{0.1867} \B\\ %
        \hline
        \hline
    \end{tabu}
    }
    }
\vspace{-0.5em}
\label{tab:embedding}
\end{table*}

\begin{table*}[t]\centering
\captionsetup[table]{skip=4pt}
\caption{Comparison of alterantive design of mask-aware pixel-shuffle down-sampling} %
    \resizebox{0.99\textwidth}{!}{
    {
    \tabulinesep=0mm
    \begin{tabular}{@{\extracolsep{3pt}}c c c c c c | c c c c c| c c c c c@{}}
        \hline\hline
        \multirow{2}{*}{Model} & \multicolumn{5}{c}{0.01\%-20\%} & \multicolumn{5}{c}{20\%-40\%} & \multicolumn{5}{c}{40\%-60\%} \T\B\\
        \cline{2-6} \cline{7-11} \cline{12-16}
          & PSNR$\uparrow$ & SSIM$\uparrow$ & L1$\downarrow$ & FID$\downarrow$ & LPIPS$\downarrow$ & PSNR$\uparrow$ & SSIM$\uparrow$ & L1$\downarrow$ & FID$\downarrow$ & LPIPS$\downarrow$ & PSNR$\uparrow$ & SSIM$\uparrow$ & L1$\downarrow$ & FID$\downarrow$ & LPIPS$\downarrow$  \T\B\\
        \hline
        CD  & 34.3159 & 0.9641 & 0.4842 & 1.4875 & 0.0380 & 26.6809 & 0.8846 & 1.6499 & 4.9362 & 0.1088 & 22.2408 & 0.7761 & 3.5828 & 8.2493 & 0.1979 \T\\  
        PD   & 34.5229 & 0.9647 & 0.4787  & 1.3729 & 0.0393 & 26.8446 & 0.8865 & 1.6328 & 4.8756 & 0.1116 & 22.4448 & 0.7795 & 3.5466 & 8.0196 & 0.2023\\  
        MPD (Ours)  & \textbf{34.5820} & \textbf{0.9649}&	\textbf{0.4733} & \textbf{1.3542} & 	\textbf{0.0375} &	\textbf{26.9327} &	\textbf{0.8867} &	\textbf{1.6089} & \textbf{4.6891} &	\textbf{0.1085} &	\textbf{22.4769} & \textbf{0.7812} & \textbf{3.5001} & \textbf{7.8697} & \textbf{0.1972} \B\\
        \hline\hline
    \end{tabular}
    }
    }
    \vspace{-0.5em}
    \label{tab:MPD_Ablation2}
\end{table*} 

\begin{table*}[h]\centering
\captionsetup[table]{skip=4pt}
\caption{Hyper-parameter tuning on the weights associated with different losses.}
    \resizebox{0.99\textwidth}{!}{
    {
    \tabulinesep=0mm
    \begin{tabular}{@{\extracolsep{3pt}}c c c c c c| c c c c c |c c c c c@{}}
        \hline\hline
        \multirow{2}{*}{Model} & \multicolumn{5}{c}{0.01\%-20\%} & \multicolumn{5}{c}{20\%-40\%} & \multicolumn{5}{c}{40\%-60\%} \T\B\\
        \cline{2-6} \cline{7-11} \cline{12-16}
          & PSNR$\uparrow$ & SSIM$\uparrow$ & L1$\downarrow$ & FID$\downarrow$ & LPIPS$\downarrow$ & PSNR$\uparrow$ & SSIM$\uparrow$ & L1$\downarrow$ & FID$\downarrow$ & LPIPS$\downarrow$ & PSNR$\uparrow$ & SSIM$\uparrow$ & L1$\downarrow$ & FID$\downarrow$ & LPIPS$\downarrow$  \T\B\\
        \hline
        Sample A  & 34.6959 & 0.9581 & 0.4417 & 1.3143 & 0.0355 & 26.8916 & 0.8358 & 1.6470 & 4.8173 & 0.1073 & 22.7519 & 0.7850 & 3.4011 & 7.9715 & 0.1902 \T\\  
        Sample B  & 33.5762 & 0.9233 & 0.4256 & 1.0158 & 0.0384 & 26.3527 & 0.8657 & 1.5419 & 4.9836 & 0.1216 & 22.4893 & 0.7754 & 3.4581 & 8.0381 & 0.1972\\  
        Sample C (ours)  & \textbf{35.0436} & \textbf{0.9671}&	\textbf{0.4489} & \textbf{1.3542} & 	\textbf{0.0345} &	\textbf{27.2954} &	\textbf{0.8924} &	\textbf{1.5363} & \textbf{4.6891} &	\textbf{0.1016} &	\textbf{22.8473} & \textbf{0.7895} & \textbf{3.3403} & \textbf{7.8697} & \textbf{0.1867} \B\\
        \hline\hline
    \end{tabular}
    }
    }
    \label{tab:Parameter}
\end{table*} 
\begin{table*}[t]
\captionsetup[table]{skip=4pt}
\centering
\caption{Ablation study of using traditional spatial self-attention in the SCAL on the $64\times64$ resolution.}
\resizebox{2\columnwidth}{!}{
{
    \tabulinesep=0mm
    \begin{tabu}{@{\extracolsep{3pt}}c c c c c c |c c c c c |c c c c c@{}}
    \hline\hline
    \multicolumn{1}{c}{\multirow{2}{*}{Model}} & 
    \multicolumn{5}{c}{0.01\%-20\%} &
    \multicolumn{5}{c}{20\%-40\%} &
    \multicolumn{5}{c}{40\%-60\%}\T\B \\
    \cline{2-6} \cline{7-11} \cline{11-16}
    
         & PSNR$\uparrow$ & SSIM$\uparrow$ & L1$\downarrow$ & FID$\downarrow$ & LPIPS$\downarrow$ & PSNR$\uparrow$ & SSIM$\uparrow$ & L1$\downarrow$ & FID$\downarrow$ & LPIPS$\downarrow$ & PSNR$\uparrow$ & SSIM$\uparrow$ & L1$\downarrow$ & FID$\downarrow$ & LPIPS$\downarrow$\T\B\\
         
        \hline\hline
        
        w SSA   & 34.4915 & 0.9515 & 0.6812 & 1.5641 & 0.0295 & 26.2375 & 0.8726 & 2.3536 & 5.0318 & 0.0720 & 21.3534 & 0.7726 & 4.3687 & 8.4753 & \textbf{0.1312}\T\\
        
        SCAL (Ours)  & \textbf{34.5849} & \textbf{0.9538}&	\textbf{0.6715} &	\textbf{1.5119} &	\textbf{0.0271} &	\textbf{26.3195} &	\textbf{0.8781} &	\textbf{2.2328} &	\textbf{4.9513} &	\textbf{0.0707} &	\textbf{21.5242} & \textbf{0.7774} & \textbf{4.2918} &	\textbf{8.4518} & \textbf{0.1312} \B\\ %
        
        \hline
        \hline
        
    \end{tabu}
    }
    }
\vspace{-1.5em}
\label{tab:spatialatt}
\end{table*}

\begin{figure}[t]
\begin{center}
\centerline{\includegraphics[scale=2.3]{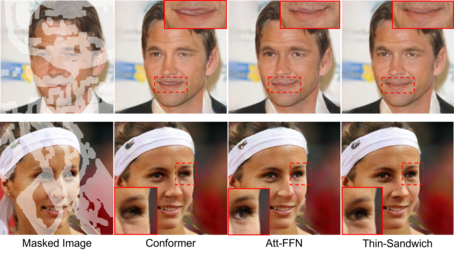}}
\end{center}
    \vspace{-2.0em}
   \caption{\revision{Visual results of the variants of sandwich.}}
\label{fig:thin-sand}
\end{figure}

\begin{figure}[t]
\begin{center}
\centerline{\includegraphics[scale=2.3]{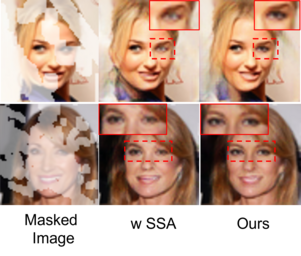}}
\end{center}
    \vspace{-2.5em}
   \caption{\revision{Visual results of the variants of SCAL.}}
   \vspace{-2.0em}
\label{fig:SSA}
\end{figure}

\subsection{Ablation Study and Parameter Analysis}\label{Ablation_study}
We conducted a series of ablation experiments on the CelebA-HQ dataset to evaluate the impact of each proposed component by downgrading them. All models are trained for 30,000 iterations. \revision{Our quantitative comparison results, which are presented in Tab.~\ref{tab:ablation}, demonstrate the effectiveness of our key contributions. \enquote{U-Net w self-attention} (model D) is the variant in which we ablate HINT to only include channel self-attention~\cite{zamir2022restormer}, a single FFN, and convolutional down-sampling. We also present visual results for a more intuitive demonstration in Fig.~\ref{fig:ablation}.}

\textbf{Spatially-activated Channel Attention Layer}
Our proposed SCAL captures channel-wise long-range dependencies while complementing the spatial attention in an efficient manner. We suggest that introducing the spatial attention identifies \enquote{where} the important regions are. As illustrated in Fig.~\ref{fig:ablation} (model C), after removing the spatial attention, the model is not confident enough to determine if an eye is missing on the left, thus generating a very blurry left eye. We substituted our spatial branch with the basic gated mechanism from \cite{deng2022t} (model E) to evaluate our superiority. \secrevision{In Tab.~\ref{tab:spatialatt}, we replace the spatial branch with traditional spatial self-attention (SSA), denoted as ‘w SSA’, to evaluate our efficiency. However, due to the significant computational cost of SSA, we have to resize the image to $64\times64$ to train on a single A100. For a fair comparison, all experiments in Tab.~\ref{tab:spatialatt} are conducted on $64\times64$ images.  We notice that the significant computational cost of SSA does not bring better performance, which is reflected in the ambiguous features with the blur texture (shown in Fig.~\ref{fig:SSA}).}

\textbf{Mask-aware Pixel-shuffle Down-sampling}
Our novel downsampling method based on pixel shuffling maintains a consistent flow of valid information within the transformer. First, to demonstrate the feasibility of pixel-shuffle down-sampling, we compare the performance of convolutional downsampling, conventional PD and the proposed MPD on the baseline. We ablate all proposed designs to build the baseline, including the ``Attention-FFN'' structure, single channel-wise self-attention branch, and conventional PD. As shown in Tab.~\ref{tab:MPD_Ablation2}, directly using conventional PD provides an overall improvement compared to convolutional down-sampling, but leads to a decline in LPIPS. We attribute this degradation to the incoherence of invalid information, which causes inaccurate transfer of high-level feature representations. MPD solves this problem and improves LPIPS significantly. 
Correspondingly, in Tab.~\ref{tab:ablation}, the performance of HINT suffers the largest drop when we replace the MPD with conventional PD. As shown in Fig.~\ref{fig:ablation} (model A), the facial attributes are severely drifting when MPD is removed. 



\textbf{Sandwich-shaped Transformer Block}\label{ablation-sandwich}
We introduce an FFN-SCAL-FFN block to effectively manage the limited flow of information. As evidenced by the results in Tab.~\ref{tab:ablation}, removing the first FFN in the sandwich leads to a notable decrease across all four metrics.  In Fig.~\ref{fig:ablation} (model B), the model fails to learn a good enough feature representation of the eyeball and nose, resulting in unclear textures for the generated left eye and nose. \secrevision{Furthermore, to confirm that the effectiveness of the proposed Sandwich Network is not merely attributed to an increase in the number of parameters, we implemented a lightweight variant that diminishes the parameter count in both Feedforward Neural Networks (FFNs) by 50\%. This thin
\enquote{Sandwich} configuration possesses an equivalent number of parameters as the \enquote{Attention-FFN} architecture. Furthermore, we substituted our \enquote{FFN-SCAL-FFN} with the Conformer structure (FFN-Attention-CONV-FFN)~\cite{gulati2020conformer} to evaluate our superiority. As shown in Fig.~\ref{fig:thin-sand}, the proposed Thin-Sandwich helps the model to learn a better feature representation of the eyeball and mouth to provide clearer texture details. Although Conformer also has a \enquote{sandwich} structure, it moves the convolutional layer that can extract local spatial feature behind the attention. Therefore, it does not embed good enough features for the following attention layer, making it difficult to generate clear texture and structure in the generated area.
}As shown in Table~\ref{tab:thin_sandwich}, the experimental results substantiate that, given an equal parameter quantity, the Sandwich module enhances the overall performance of the model.


\textbf{Decision for the Last Skip Connection}\label{skip}
To harness the low-level texture and structural features derived from the encoder, we refrain from utilising $1 \times 1$ convolution for modulating the number of channels post the last skip connection. The contrast between the two strategies is enumerated in Tab.~\ref{tab:last-skip-connect}.

\textbf{Embedding Layer}\label{embedding}
In the embedding layer, we adopt a gated convolutional layer with padding to embed the input without downsampling. In contrast to prior works using $7\times7$ convolutional layers to project the input~\cite{guo2021image,li2022misf,nazeri2019edgeconnect}, a smaller kernel size ($3\times3$) is employed in our embedding layer to obtain more fine-grained features.
As illustrated in Table~\ref{tab:embedding}, the smaller kernel gains better performance.

\textbf{Parameter Tuning}\label{Parameter_tuning}
To tune HINT, we employ Optuna~\cite{akiba2019optuna} to identify the best set of hyper-parameters in terms of different values of weights of our loss components. The top three sets of combinations are $\lambda_i$: sample A [1,1,0.5,2], sample B [1,60,1,2], sample C [1,250,0.1,0.001], as shown in Tab. ~\ref{tab:Parameter}. We implement the sample C for all of the experiments.

\section{Conclusion and Discussions}
\secrevision{We propose HINT, an end-to-end Transformer for image inpainting with the proposed MPD module to ensure information remains intact and consistent throughout the encoding process. The MPD is a plug-and-play module, which is easy to adopt to the other multimedia tasks that require masking process, such as video edit, and animation edit. Our SCAL, enhanced by the proposed \enquote{sandwich} module, captures long-range dependencies while remaining spatial awareness, to boosting the capacity of representation learning in a cheap approach, which could potentially benefit multimedia tasks that are based on channel self-attention.}


The proposed components contribute to each other and drive HINT to recover high-quality completed images. Experimental results demonstrate that HINT overall surpasses the current state of the art on four datasets~\cite{liu2015deep,karras2017progressive,zhou2017places,yu2019dunhuang}, with particularly notable improvements observed on facial datasets~\cite{liu2015deep,karras2017progressive}. Extensive qualitative evaluations demonstrate the superior image quality achieved by our framework.

As a direction of future research, HINT can be improved by employing geometric information~\cite{nazeri2019edgeconnect,chen22feasibility} by simply adding an indicator or incorporating a multi-task architecture, to get better structural consistency. Furthermore, considering the success of existing work~\cite{Ni_2023_CVPR}, HINT can be potentially upgraded to a text-guided image inpainting system by introducing the pre-trained multi-model features to interpret the text feature into the latent space.

Furthermore, unlike existing multi-step approaches ~\cite{zheng2022bridging, li2022mat, wan2021high}, as HINT is already able to recover high-quality completed images without requiring additional refinement process, a second stage of reconstruction could further enhance the quality of the results. Constrained by the current limited computing resources, we will implement another refinement network in the second step, utilizing the results from HINT as inputs and fine-tuning them in the same scale. Two networks are trained separately, thereby avoiding the large number of parameters introduced by joint training.

\section*{Acknowledgements}
This research is supported in part by the EPSRC NortHFutures project (ref: EP/X031012/1).

\bibliographystyle{IEEEtran}
\bibliography{egbib.bib}

\begin{thebibliography}{10}
\providecommand{\url}[1]{#1}
\csname url@samestyle\endcsname
\providecommand{\newblock}{\relax}
\providecommand{\bibinfo}[2]{#2}
\providecommand{\BIBentrySTDinterwordspacing}{\spaceskip=0pt\relax}
\providecommand{\BIBentryALTinterwordstretchfactor}{4}
\providecommand{\BIBentryALTinterwordspacing}{\spaceskip=\fontdimen2\font plus
\BIBentryALTinterwordstretchfactor\fontdimen3\font minus \fontdimen4\font\relax}
\providecommand{\BIBforeignlanguage}[2]{{%
\expandafter\ifx\csname l@#1\endcsname\relax
\typeout{** WARNING: IEEEtran.bst: No hyphenation pattern has been}%
\typeout{** loaded for the language `#1'. Using the pattern for}%
\typeout{** the default language instead.}%
\else
\language=\csname l@#1\endcsname
\fi
#2}}
\providecommand{\BIBdecl}{\relax}
\BIBdecl

\bibitem{jo2019sc}
Y.~Jo and J.~Park, ``Sc-fegan: Face editing generative adversarial network with user's sketch and color,'' in \emph{Proceedings of the IEEE/CVF international conference on computer vision}, 2019, pp. 1745--1753.

\bibitem{wei2019shadow}
J.~Wei, C.~Long, H.~Zou, and C.~Xiao, ``Shadow inpainting and removal using generative adversarial networks with slice convolutions,'' in \emph{Computer Graphics Forum}, vol.~38, no.~7.\hskip 1em plus 0.5em minus 0.4em\relax Wiley Online Library, 2019, pp. 381--392.

\bibitem{atapour2019dealing}
A.~Atapour-Abarghouei and T.~P. Breckon, ``Dealing with missing depth: recent advances in depth image completion and estimation,'' \emph{RGB-D Image analysis and processing}, pp. 15--50, 2019.

\bibitem{pathak2016context}
D.~Pathak, P.~Krahenbuhl, J.~Donahue, T.~Darrell, and A.~A. Efros, ``Context encoders: Feature learning by inpainting,'' in \emph{Proceedings of the IEEE conference on computer vision and pattern recognition}, 2016, pp. 2536--2544.

\bibitem{iizuka2017globally}
S.~Iizuka, E.~Simo-Serra, and H.~Ishikawa, ``Globally and locally consistent image completion,'' \emph{ACM Transactions on Graphics (ToG)}, vol.~36, no.~4, pp. 1--14, 2017.

\bibitem{liu2018image}
G.~Liu, F.~A. Reda, K.~J. Shih, T.-C. Wang, A.~Tao, and B.~Catanzaro, ``Image inpainting for irregular holes using partial convolutions,'' in \emph{Proceedings of the European conference on computer vision (ECCV)}, 2018, pp. 85--100.

\bibitem{yu2019free}
J.~Yu, Z.~Lin, J.~Yang, X.~Shen, X.~Lu, and T.~S. Huang, ``Free-form image inpainting with gated convolution,'' in \emph{Proceedings of the IEEE/CVF International Conference on Computer Vision}, 2019, pp. 4471--4480.

\bibitem{cao2021learning}
C.~Cao and Y.~Fu, ``Learning a sketch tensor space for image inpainting of man-made scenes,'' in \emph{Proceedings of the IEEE/CVF International Conference on Computer Vision}, 2021, pp. 14\,509--14\,518.

\bibitem{yu2018generative}
J.~Yu, Z.~Lin, J.~Yang, X.~Shen, X.~Lu, and T.~S. Huang, ``Generative image inpainting with contextual attention,'' in \emph{Proceedings of the IEEE conference on computer vision and pattern recognition}, 2018, pp. 5505--5514.

\bibitem{guo2021image}
X.~Guo, H.~Yang, and D.~Huang, ``Image inpainting via conditional texture and structure dual generation,'' in \emph{Proceedings of the IEEE/CVF International Conference on Computer Vision}, 2021, pp. 14\,134--14\,143.

\bibitem{wang2019musical}
N.~Wang, J.~Li, L.~Zhang, and B.~Du, ``Musical: Multi-scale image contextual attention learning for inpainting.'' in \emph{IJCAI}, 2019, pp. 3748--3754.

\bibitem{sun2022learning}
H.~Sun, W.~Li, Y.~Duan, J.~Zhou, and J.~Lu, ``Learning adaptive patch generators for mask-robust image inpainting,'' \emph{IEEE Transactions on Multimedia}, 2022.

\bibitem{sridevi2019image}
G.~Sridevi and S.~Srinivas~Kumar, ``Image inpainting based on fractional-order nonlinear diffusion for image reconstruction,'' \emph{Circuits, Systems, and Signal Processing}, vol.~38, pp. 3802--3817, 2019.

\bibitem{atapour2016back}
A.~Atapour-Abarghouei, G.~P. de~La~Garanderie, and T.~P. Breckon, ``Back to butterworth-a fourier basis for 3d surface relief hole filling within rgb-d imagery,'' in \emph{2016 23rd International Conference on Pattern Recognition (ICPR)}.\hskip 1em plus 0.5em minus 0.4em\relax IEEE, 2016, pp. 2813--2818.

\bibitem{barnes2009patchmatch}
C.~Barnes, E.~Shechtman, A.~Finkelstein, and D.~B. Goldman, ``Patchmatch: A randomized correspondence algorithm for structural image editing,'' \emph{ACM Trans. Graph.}, vol.~28, no.~3, p.~24, 2009.

\bibitem{zhang2022w}
R.~Zhang, W.~Quan, Y.~Zhang, J.~Wang, and D.-M. Yan, ``W-net: Structure and texture interaction for image inpainting,'' \emph{IEEE Transactions on Multimedia}, 2022.

\bibitem{zhang2022pluralistic}
Y.~Zhang, X.~Zhang, C.~Shi, X.~Wu, X.~Li, J.~Peng, K.~Cao, J.~Lv, and J.~Zhou, ``Pluralistic face inpainting with transformation of attribute information,'' \emph{IEEE Transactions on Multimedia}, 2022.

\bibitem{nazeri2019edgeconnect}
K.~Nazeri, E.~Ng, T.~Joseph, F.~Z. Qureshi, and M.~Ebrahimi, ``Edgeconnect: Generative image inpainting with adversarial edge learning,'' \emph{arXiv preprint arXiv:1901.00212}, 2019.

\bibitem{uddin2020global}
S.~Uddin and Y.~J. Jung, ``Global and local attention-based free-form image inpainting,'' \emph{Sensors}, vol.~20, no.~11, p. 3204, 2020.

\bibitem{zheng2022bridging}
C.~Zheng, T.-J. Cham, J.~Cai, and D.~Phung, ``Bridging global context interactions for high-fidelity image completion,'' in \emph{Proceedings of the IEEE/CVF Conference on Computer Vision and Pattern Recognition}, 2022, pp. 11\,512--11\,522.

\bibitem{wan2021high}
Z.~Wan, J.~Zhang, D.~Chen, and J.~Liao, ``High-fidelity pluralistic image completion with transformers,'' in \emph{Proceedings of the IEEE/CVF International Conference on Computer Vision}, 2021, pp. 4692--4701.

\bibitem{li2022mat}
W.~Li, Z.~Lin, K.~Zhou, L.~Qi, Y.~Wang, and J.~Jia, ``Mat: Mask-aware transformer for large hole image inpainting,'' in \emph{Proceedings of the IEEE/CVF conference on computer vision and pattern recognition}, 2022, pp. 10\,758--10\,768.

\bibitem{suvorov2022resolution}
R.~Suvorov, E.~Logacheva, A.~Mashikhin, A.~Remizova, A.~Ashukha, A.~Silvestrov, N.~Kong, H.~Goka, K.~Park, and V.~Lempitsky, ``Resolution-robust large mask inpainting with fourier convolutions,'' in \emph{Proceedings of the IEEE/CVF winter conference on applications of computer vision}, 2022, pp. 2149--2159.

\bibitem{li2022misf}
X.~Li, Q.~Guo, D.~Lin, P.~Li, W.~Feng, and S.~Wang, ``Misf: Multi-level interactive siamese filtering for high-fidelity image inpainting,'' in \emph{Proceedings of the IEEE/CVF Conference on Computer Vision and Pattern Recognition}, 2022, pp. 1869--1878.

\bibitem{guo2021jpgnet}
Q.~Guo, X.~Li, F.~Juefei-Xu, H.~Yu, Y.~Liu, and S.~Wang, ``Jpgnet: Joint predictive filtering and generative network for image inpainting,'' in \emph{Proceedings of the 29th ACM International Conference on Multimedia}, 2021, pp. 386--394.

\bibitem{karras2017progressive}
T.~Karras, T.~Aila, S.~Laine, and J.~Lehtinen, ``Progressive growing of gans for improved quality, stability, and variation,'' \emph{arXiv preprint arXiv:1710.10196}, 2017.

\bibitem{zhou2017places}
B.~Zhou, A.~Lapedriza, A.~Khosla, A.~Oliva, and A.~Torralba, ``Places: A 10 million image database for scene recognition,'' \emph{IEEE transactions on pattern analysis and machine intelligence}, vol.~40, no.~6, pp. 1452--1464, 2017.

\bibitem{yu2019dunhuang}
T.~Yu, S.~Zhang, C.~Lin, S.~You, J.~Wu, J.~Zhang, X.~Ding, and H.~An, ``Dunhuang grottoes painting dataset and benchmark,'' \emph{arXiv preprint arXiv:1907.04589}, 2019.

\bibitem{zhao2017random}
G.~Zhao, J.~Wang, Z.~Zhang \emph{et~al.}, ``Random shifting for cnn: a solution to reduce information loss in down-sampling layers.'' in \emph{IJCAI}, 2017, pp. 3476--3482.

\bibitem{a2021beyond}
S.~A~Sharif, R.~A. Naqvi, and M.~Biswas, ``Beyond joint demosaicking and denoising: An image processing pipeline for a pixel-bin image sensor,'' in \emph{Proceedings of the IEEE/CVF Conference on Computer Vision and Pattern Recognition}, 2021, pp. 233--242.

\bibitem{yue2021semi}
Z.~Yue, J.~Xie, Q.~Zhao, and D.~Meng, ``Semi-supervised video deraining with dynamical rain generator,'' in \emph{Proceedings of the IEEE/CVF Conference on Computer Vision and Pattern Recognition}, 2021, pp. 642--652.

\bibitem{wang2021learning}
L.~Wang, Y.~Wang, Z.~Lin, J.~Yang, W.~An, and Y.~Guo, ``Learning a single network for scale-arbitrary super-resolution,'' in \emph{Proceedings of the IEEE/CVF international conference on computer vision}, 2021, pp. 4801--4810.

\bibitem{zhou2020awgn}
Y.~Zhou, J.~Jiao, H.~Huang, Y.~Wang, J.~Wang, H.~Shi, and T.~Huang, ``When awgn-based denoiser meets real noises,'' in \emph{Proceedings of the AAAI Conference on Artificial Intelligence}, vol.~34, no.~07, 2020, pp. 13\,074--13\,081.

\bibitem{zamir2022restormer}
S.~W. Zamir, A.~Arora, S.~Khan, M.~Hayat, F.~S. Khan, and M.-H. Yang, ``Restormer: Efficient transformer for high-resolution image restoration,'' in \emph{Proceedings of the IEEE/CVF Conference on Computer Vision and Pattern Recognition}, 2022, pp. 5728--5739.

\bibitem{69}
M.~Li, Y.~Fu, and Y.~Zhang, ``Spatial-spectral transformer for hyperspectral image denoising,'' in \emph{Proceedings of the AAAI Conference on Artificial Intelligence}, vol.~37, no.~1, 2023, pp. 1368--1376.

\bibitem{70}
S.-I. Jang, T.~Pan, Y.~Li, P.~Heidari, J.~Chen, Q.~Li, and K.~Gong, ``Spach transformer: spatial and channel-wise transformer based on local and global self-attentions for pet image denoising,'' \emph{IEEE Transactions on Medical Imaging}, 2023.

\bibitem{71}
L.~Wang, M.~Cao, Y.~Zhong, and X.~Yuan, ``Spatial-temporal transformer for video snapshot compressive imaging,'' \emph{IEEE Transactions on Pattern Analysis and Machine Intelligence}, 2022.

\bibitem{liu2015deep}
Z.~Liu, P.~Luo, X.~Wang, and X.~Tang, ``Deep learning face attributes in the wild,'' in \emph{Proceedings of the IEEE international conference on computer vision}, 2015, pp. 3730--3738.

\bibitem{goodfellow2014generative}
I.~J. Goodfellow, J.~Pouget-Abadie, M.~Mirza, B.~Xu, D.~Warde-Farley, S.~Ozair, A.~Courville, and Y.~Bengio, ``Generative adversarial networks,'' \emph{arXiv e-prints}, pp. arXiv--1406, 2014.

\bibitem{li2020recurrent}
J.~Li, N.~Wang, L.~Zhang, B.~Du, and D.~Tao, ``Recurrent feature reasoning for image inpainting,'' in \emph{Proceedings of the IEEE/CVF Conference on Computer Vision and Pattern Recognition}, 2020, pp. 7760--7768.

\bibitem{peng2021generating}
J.~Peng, D.~Liu, S.~Xu, and H.~Li, ``Generating diverse structure for image inpainting with hierarchical vq-vae,'' in \emph{Proceedings of the IEEE/CVF Conference on Computer Vision and Pattern Recognition}, 2021, pp. 10\,775--10\,784.

\bibitem{wu2021deep}
H.~Wu, J.~Zhou, and Y.~Li, ``Deep generative model for image inpainting with local binary pattern learning and spatial attention,'' \emph{IEEE Transactions on Multimedia}, vol.~24, pp. 4016--4027, 2021.

\bibitem{chen2023inclg}
S.~Chen, A.~Atapour-Abarghouei, E.~S. Ho, and H.~P. Shum, ``Inclg: Inpainting for non-cleft lip generation with a multi-task image processing network,'' \emph{Software Impacts}, vol.~17, p. 100517, 2023.

\bibitem{yu2020region}
T.~Yu, Z.~Guo, X.~Jin, S.~Wu, Z.~Chen, W.~Li, Z.~Zhang, and S.~Liu, ``Region normalization for image inpainting,'' in \emph{Proceedings of the AAAI Conference on Artificial Intelligence}, vol.~34, no.~07, 2020, pp. 12\,733--12\,740.

\bibitem{76}
H.~Zheng, Z.~Lin, J.~Lu, S.~Cohen, E.~Shechtman, C.~Barnes, J.~Zhang, N.~Xu, S.~Amirghodsi, and J.~Luo, ``Image inpainting with cascaded modulation gan and object-aware training,'' in \emph{European Conference on Computer Vision}.\hskip 1em plus 0.5em minus 0.4em\relax Springer, 2022, pp. 277--296.

\bibitem{vaswani2017attention}
A.~Vaswani, N.~Shazeer, N.~Parmar, J.~Uszkoreit, L.~Jones, A.~N. Gomez, {\L}.~Kaiser, and I.~Polosukhin, ``Attention is all you need,'' \emph{Advances in neural information processing systems}, vol.~30, 2017.

\bibitem{liu2021swin}
Z.~Liu, Y.~Lin, Y.~Cao, H.~Hu, Y.~Wei, Z.~Zhang, S.~Lin, and B.~Guo, ``Swin transformer: Hierarchical vision transformer using shifted windows,'' in \emph{Proceedings of the IEEE/CVF international conference on computer vision}, 2021, pp. 10\,012--10\,022.

\bibitem{dosovitskiy2020image}
A.~Dosovitskiy, L.~Beyer, A.~Kolesnikov, D.~Weissenborn, X.~Zhai, T.~Unterthiner, M.~Dehghani, M.~Minderer, G.~Heigold, S.~Gelly \emph{et~al.}, ``An image is worth 16x16 words: Transformers for image recognition at scale,'' \emph{arXiv preprint arXiv:2010.11929}, 2020.

\bibitem{yu2021diverse}
Y.~Yu, F.~Zhan, R.~Wu, J.~Pan, K.~Cui, S.~Lu, F.~Ma, X.~Xie, and C.~Miao, ``Diverse image inpainting with bidirectional and autoregressive transformers,'' in \emph{Proceedings of the 29th ACM International Conference on Multimedia}, 2021, pp. 69--78.

\bibitem{zhang2023mutual}
Y.~Zhang, Y.~Liu, R.~Hu, Q.~Wu, and J.~Zhang, ``Mutual dual-task generator with adaptive attention fusion for image inpainting,'' \emph{IEEE Transactions on Multimedia}, 2023.

\bibitem{deng2021learning}
Y.~Deng, S.~Hui, S.~Zhou, D.~Meng, and J.~Wang, ``Learning contextual transformer network for image inpainting,'' in \emph{Proceedings of the 29th ACM International Conference on Multimedia}, 2021, pp. 2529--2538.

\bibitem{deng2022t}
------, ``T-former: An efficient transformer for image inpainting,'' in \emph{Proceedings of the 30th ACM International Conference on Multimedia}, 2022, pp. 6559--6568.

\bibitem{hendrycks2016gaussian}
D.~Hendrycks and K.~Gimpel, ``Gaussian error linear units (gelus),'' \emph{arXiv preprint arXiv:1606.08415}, 2016.

\bibitem{xiao2021early}
T.~Xiao, M.~Singh, E.~Mintun, T.~Darrell, P.~Doll{\'a}r, and R.~Girshick, ``Early convolutions help transformers see better,'' \emph{Advances in Neural Information Processing Systems}, vol.~34, pp. 30\,392--30\,400, 2021.

\bibitem{shi2016real}
W.~Shi, J.~Caballero, F.~Husz{\'a}r, J.~Totz, A.~P. Aitken, R.~Bishop, D.~Rueckert, and Z.~Wang, ``Real-time single image and video super-resolution using an efficient sub-pixel convolutional neural network,'' in \emph{Proceedings of the IEEE conference on computer vision and pattern recognition}, 2016, pp. 1874--1883.

\bibitem{chollet2017xception}
F.~Chollet, ``Xception: Deep learning with depthwise separable convolutions,'' in \emph{Proceedings of the IEEE conference on computer vision and pattern recognition}, 2017, pp. 1251--1258.

\bibitem{fu2019dual}
J.~Fu, J.~Liu, H.~Tian, Y.~Li, Y.~Bao, Z.~Fang, and H.~Lu, ``Dual attention network for scene segmentation,'' in \emph{Proceedings of the IEEE/CVF conference on computer vision and pattern recognition}, 2019, pp. 3146--3154.

\bibitem{fukushima1975cognitron}
K.~Fukushima, ``Cognitron: A self-organizing multilayered neural network,'' \emph{Biological cybernetics}, vol.~20, no. 3-4, pp. 121--136, 1975.

\bibitem{park2019specaugment}
D.~S. Park, W.~Chan, Y.~Zhang, C.-C. Chiu, B.~Zoph, E.~D. Cubuk, and Q.~V. Le, ``Specaugment: A simple data augmentation method for automatic speech recognition,'' \emph{Proc. Interspeech 2019}, pp. 2613--2617, 2019.

\bibitem{park2020specaugment}
D.~S. Park, Y.~Zhang, C.-C. Chiu, Y.~Chen, B.~Li, W.~Chan, Q.~V. Le, and Y.~Wu, ``Specaugment on large scale datasets,'' in \emph{ICASSP 2020-2020 IEEE International Conference on Acoustics, Speech and Signal Processing (ICASSP)}.\hskip 1em plus 0.5em minus 0.4em\relax IEEE, 2020, pp. 6879--6883.

\bibitem{gulati2020conformer}
A.~Gulati, J.~Qin, C.-C. Chiu, N.~Parmar, Y.~Zhang, J.~Yu, W.~Han, S.~Wang, Z.~Zhang, Y.~Wu \emph{et~al.}, ``Conformer: Convolution-augmented transformer for speech recognition,'' \emph{arXiv preprint arXiv:2005.08100}, 2020.

\bibitem{Rombach_2022_CVPR}
R.~Rombach, A.~Blattmann, D.~Lorenz, P.~Esser, and B.~Ommer, ``High-resolution image synthesis with latent diffusion models,'' in \emph{Proceedings of the IEEE/CVF Conference on Computer Vision and Pattern Recognition}, June 2022, pp. 10\,684--10\,695.

\bibitem{yu2021wavefill}
Y.~Yu, F.~Zhan, S.~Lu, J.~Pan, F.~Ma, X.~Xie, and C.~Miao, ``Wavefill: A wavelet-based generation network for image inpainting,'' in \emph{Proceedings of the IEEE/CVF international conference on computer vision}, 2021, pp. 14\,114--14\,123.

\bibitem{Lugmayr_2022_CVPR}
A.~Lugmayr, M.~Danelljan, A.~Romero, F.~Yu, R.~Timofte, and L.~Van~Gool, ``Repaint: Inpainting using denoising diffusion probabilistic models,'' in \emph{Proceedings of the IEEE/CVF Conference on Computer Vision and Pattern Recognition}, June 2022, pp. 11\,461--11\,471.

\bibitem{ren2019structureflow}
Y.~Ren, X.~Yu, R.~Zhang, T.~H. Li, S.~Liu, and G.~Li, ``Structureflow: Image inpainting via structure-aware appearance flow,'' in \emph{Proceedings of the IEEE/CVF International Conference on Computer Vision}, 2019, pp. 181--190.

\bibitem{kingma2020method}
D.~P. Kingma, J.~A. Ba, and J.~Adam, ``A method for stochastic optimization. arxiv 2014,'' \emph{arXiv preprint arXiv:1412.6980}, vol. 106, 2020.

\bibitem{zeng2019learning}
Y.~Zeng, J.~Fu, H.~Chao, and B.~Guo, ``Learning pyramid-context encoder network for high-quality image inpainting,'' in \emph{Proceedings of the IEEE/CVF Conference on Computer Vision and Pattern Recognition}, 2019, pp. 1486--1494.

\bibitem{wang2004image}
Z.~Wang, A.~C. Bovik, H.~R. Sheikh, and E.~P. Simoncelli, ``Image quality assessment: from error visibility to structural similarity,'' \emph{IEEE transactions on image processing}, vol.~13, no.~4, pp. 600--612, 2004.

\bibitem{zhang2018unreasonable}
R.~Zhang, P.~Isola, A.~A. Efros, E.~Shechtman, and O.~Wang, ``The unreasonable effectiveness of deep features as a perceptual metric,'' in \emph{Proceedings of the IEEE conference on computer vision and pattern recognition}, 2018, pp. 586--595.

\bibitem{chang23design}
\BIBentryALTinterwordspacing
Z.~Chang, G.~A. Koulieris, and H.~P.~H. Shum, ``On the design fundamentals of diffusion models: A survey,'' \emph{arXiv}, 2023. [Online]. Available: \url{http://arxiv.org/abs/2306.04542}
\BIBentrySTDinterwordspacing

\bibitem{akiba2019optuna}
T.~Akiba, S.~Sano, T.~Yanase, T.~Ohta, and M.~Koyama, ``Optuna: A next-generation hyperparameter optimization framework,'' in \emph{Proceedings of the 25th ACM SIGKDD international conference on knowledge discovery \& data mining}, 2019, pp. 2623--2631.

\bibitem{chen22feasibility}
S.~Chen, A.~Atapour-Abarghouei, J.~Kerby, E.~S.~L. Ho, D.~C.~G. Sainsbury, S.~Butterworth, and H.~P.~H. Shum, ``A feasibility study on image inpainting for non-cleft lip generation from patients with cleft lip,'' in \emph{Proceedings of the 2022 IEEE-EMBS International Conference on Biomedical and Health Informatics}, ser. BHI '22.\hskip 1em plus 0.5em minus 0.4em\relax IEEE, 9 2022, pp. 1--4.

\bibitem{Ni_2023_CVPR}
M.~Ni, X.~Li, and W.~Zuo, ``Nuwa-lip: Language-guided image inpainting with defect-free vqgan,'' in \emph{Proceedings of the IEEE/CVF Conference on Computer Vision and Pattern Recognition (CVPR)}, June 2023, pp. 14\,183--14\,192.

\end{thebibliography}

\vspace{-4.0em}
\begin{IEEEbiography}
[{\includegraphics[width=1in,height=1.25in,clip,keepaspectratio]{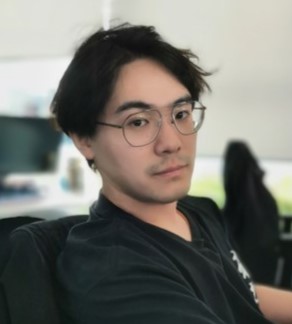}}]{Shuang Chen}
 received his B.S. degree in Automation from Shandong University of Technology in 2017, and MSc. degree in Computer Vision, Machine Learning and Robotics from University of Surrey in 2020. He is currently pursuing his PhD degree at the Department of Computer Science at Durham University in the UK. His research interests span across Computer Vision, Machine Learning, Image Processing and Image Inpainting.
\end{IEEEbiography}

\vspace{-4.0em}
\begin{IEEEbiography}
[{\includegraphics[width=1in,height=1.25in,clip,keepaspectratio]{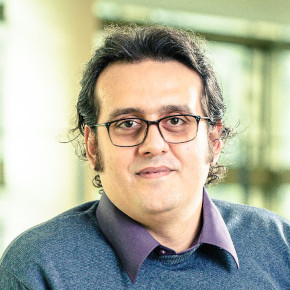}}]{Amir Atapour-Abarghouei}
is an Assistant Professor at the Department of Computer Science at Durham University. Prior to his current post, Amir held a lectureship position at Newcastle University in the UK. He received his PhD from Durham University. His primary research interests span across Computer Vision, Machine Learning and Natural Language Processing, Anomaly Detection, Brain-Computer Interface and Graph Analysis. His work includes the generalised high-impact GANomaly anomaly detection approach, which is now a part of Intel’s AI product line and used as the underlying method for anomaly detection in numerous international patents. Amir has co-organised the CVPR-NAS workshop as well as workshops at BigData (BDA4CID and BDA4HM).
\end{IEEEbiography}

\vspace{-3em}
\begin{IEEEbiography}
[{\includegraphics[width=1in,height=1.25in,clip,keepaspectratio]{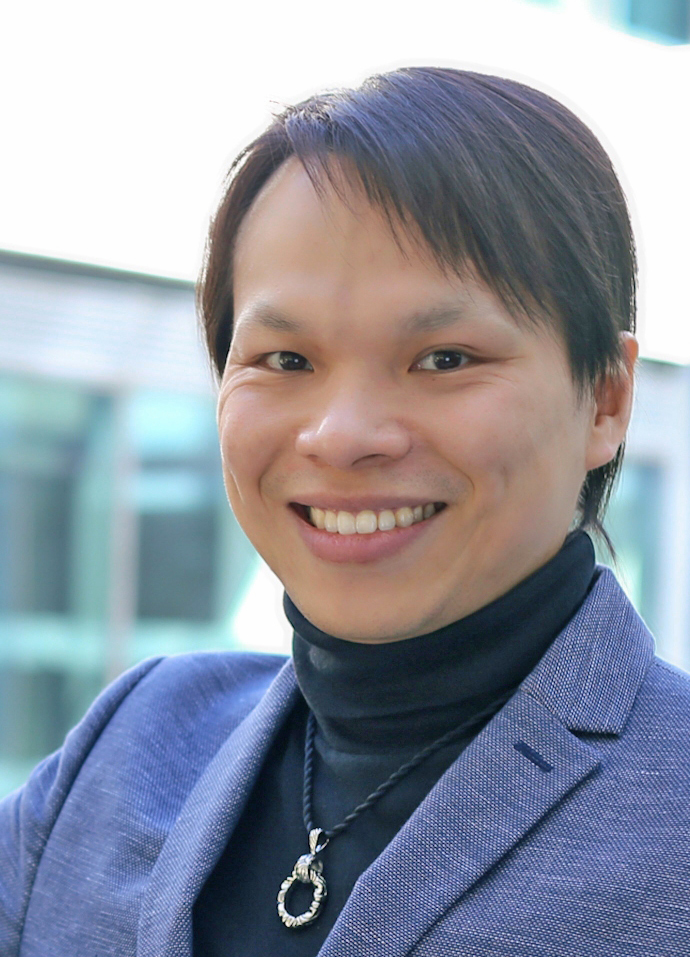}}]{Hubert P. H. Shum}
(Senior Member, IEEE) is an Associate Professor in Visual Computing and the Deputy Director of Research of the Department of Computer Science at Durham University, specialised in Spatio-Temporal Modelling and Responsible AI. He is also a Co-Founder of the Responsible Space Innovation Centre. Before this, he was an Associate Professor at Northumbria University and a Postdoctoral Researcher at RIKEN Japan. He received his PhD degree from the University of Edinburgh. He chaired conferences such as Pacific Graphics, BMVC and SCA, and has authored over 150 research publications.
\end{IEEEbiography}

\includepdf[pages=-]{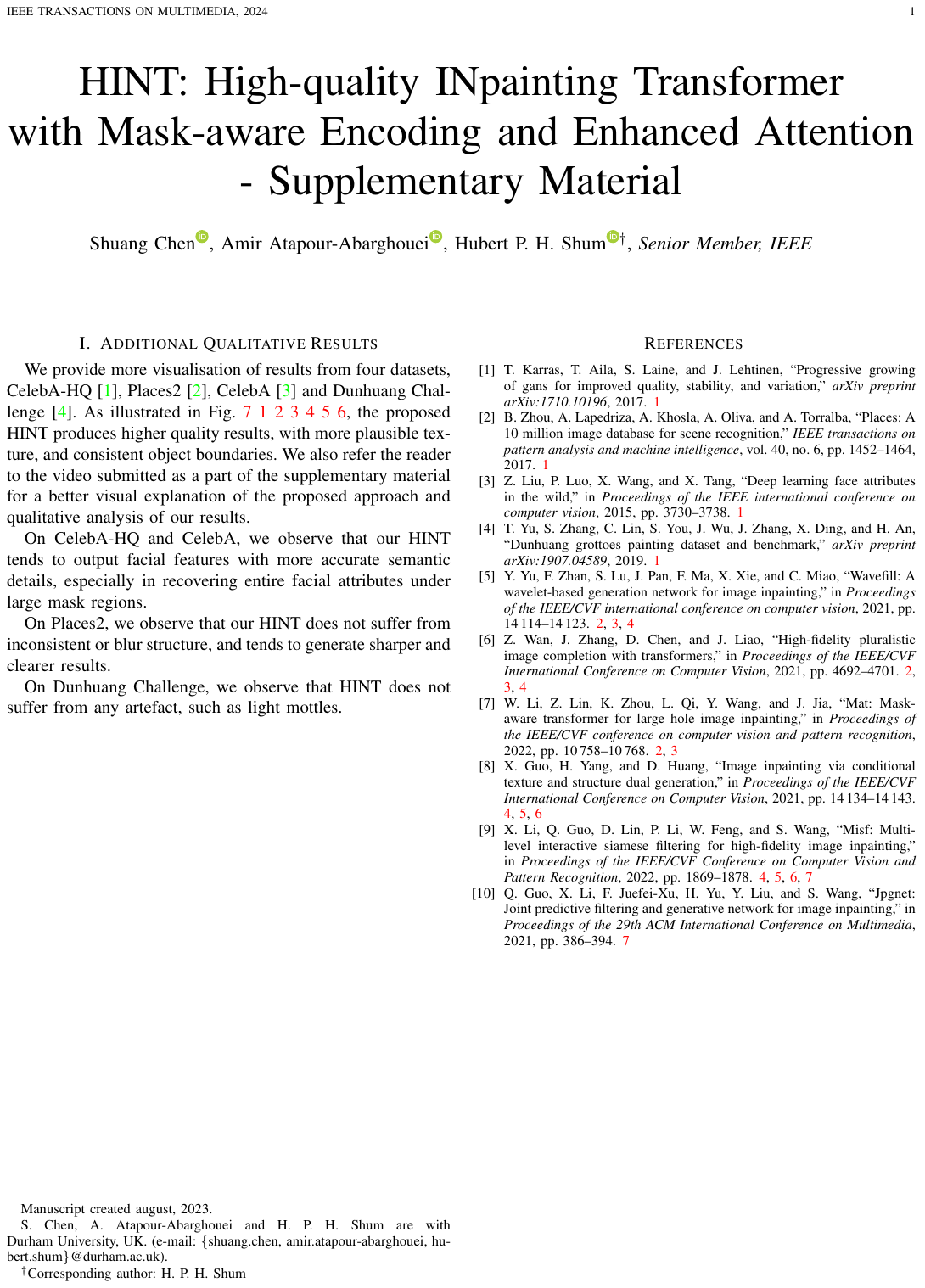}
\end{document}